\documentclass{article}

\usepackage{arxiv}

\usepackage[utf8]{inputenc} 
\usepackage[T1]{fontenc}    
\usepackage{hyperref}       
\usepackage{url}            
\usepackage{booktabs}       
\usepackage{amsfonts}       
\usepackage{nicefrac}       
\usepackage{microtype}      
\usepackage{lipsum}
\usepackage{graphicx}
\graphicspath{ {./images/} }
\usepackage{amsmath,amsfonts}
\usepackage{algorithm}
\usepackage{array}
\usepackage[caption=false,font=normalsize,labelfont=sf,textfont=sf]{subfig}
\usepackage{textcomp}
\usepackage{stfloats}
\usepackage{url}
\usepackage{multirow}

\usepackage{verbatim}
\usepackage{graphicx}
\usepackage{cite}
\usepackage{algpseudocode}

\usepackage{graphicx}
\usepackage{newtxmath}
\usepackage{subcaption}
\usepackage{orcidlink}

\title{Domain-Aware Hyperdimensional Computing for Edge Smart
Manufacturing}

\author{
 Fardin Jalil Piran \\
 School of Mechanical, Aerospace, \\ and Manufacturing Engineering\\
 University of Connecticut\\
 Storrs, CT 06269 \\
  \texttt{fardin.jalil\_piran@uconn.edu} \\
   \And
Anandkumar Patel \\
 Department of Mechanical \\ and Aerospace Engineering\\
 Rutgers University\\
 Piscataway, NJ 08854 \\
  \texttt{ap2505@scarletmail.rutgers.edu} \\
   \And
Rajiv Malhotra \\
 Department of Mechanical \\ and Aerospace Engineering\\
 Rutgers University\\
 Piscataway, NJ 08854 \\
  \texttt{rajiv.malhotra@rutgers.edu} \\
  \And
 Farhad Imani \\
 School of Mechanical, Aerospace, \\ and Manufacturing Engineering\\
 University of Connecticut\\
 Storrs, CT 06269 \\
  \texttt{farhad.imani@uconn.edu} \\
}

\begin{document}
\maketitle
\begin{abstract}
Smart manufacturing requires on-device intelligence that meets strict latency and energy budgets. \emph{HyperDimensional Computing (HDC)} offers a lightweight alternative by encoding data as high-dimensional hypervectors and computing with simple operations. Prior studies often assume that the qualitative relation between HDC hyperparameters and performance is stable across applications. Our analysis of two representative tasks, signal-based quality monitoring in Computer Numerical Control (CNC) machining and image-based defect detection in Laser Powder Bed Fusion (LPBF), shows that this assumption does not hold. We map how encoder type, projection variance, hypervector dimensionality, and data regime shape accuracy, inference latency, training time, and training energy. A formal complexity model explains predictable trends in encoding and similarity computation and reveals nonmonotonic interactions with retraining that preclude a closed-form optimum. Empirically, signals favor nonlinear Random Fourier Features with more exclusive encodings and saturate in accuracy beyond moderate dimensionality. Images favor linear Random Projection, achieve high accuracy with small dimensionality, and depend more on sample count than on dimensionality. Guided by these insights, we tune HDC under multiobjective constraints that reflect edge deployment and obtain models that match or exceed the accuracy of state-of-the-art deep learning and Transformer models while delivering at least \(6\times\) faster inference and more than \(40\times\) lower training energy. These results demonstrate that domain-aware HDC encoding is necessary and that tuned HDC offers a practical, scalable path to real-time industrial AI on constrained hardware. Future work will enable adaptive encoder and hyperparameter selection, expand evaluation to additional manufacturing modalities, and validate on low-power accelerators.
\\
\\
\end{abstract}


\section{Introduction}
\label{sec:introduction}

Integrating digital intelligence with physical manufacturing processes offers significant economic and environmental advantages, such as reducing waste, improving efficiency, and increasing equipment utilization~\cite{matsunaga2022optimization,sahoo2022smart}. Achieving these benefits requires deploying Artificial Intelligence (AI) models directly on edge devices to support latency-sensitive tasks, such as anomaly detection, predictive maintenance, and closed-loop control~\cite{gao2022guest,chentransfer,cleeman2025scalable}. Unlike cloud-based systems, edge computing performs data analytics close to the source, reducing transmission overhead, conserving bandwidth, and enhancing security and resilience~\cite{zhang2021edge,ray2021sdn,modupe2024reviewing}. This capability is particularly valuable in manufacturing, where real-time analysis of sensor streams enables fast anomaly detection and autonomous decision-making~\cite{piran2025privacy}. However, conventional deep learning models remain computationally intensive, energy-hungry, and often impractical for resource-constrained enterprises~\cite{han2015deep,dequino2025optimizing}.  

\emph{HyperDimensional Computing (HDC)} provides a lightweight alternative to deep learning by encoding data into high-dimensional vectors, or \emph{hypervectors}, and manipulating them through simple element-wise operations such as binding, bundling, and permutation~\cite{aygun2023learning,masum2025ams}. Since hypervectors can be stored in low-precision memory and processed with bitwise operations, HDC naturally aligns with hardware acceleration and energy-efficient computation~\cite{desislavov2023trends}. Prior studies have explored binarization, dimensionality reduction, iterative co-tuning, and hardware-aware methods to lower HDC cost while maintaining accuracy~\cite{xu2025hypermetric,ponzina2024microhd,kazemi2021mimhd}. A prevailing assumption in this literature is that while the exact numerical values of optimal parameters may vary across datasets, the \emph{qualitative shape} of performance trade-offs remains invariant. For example, accuracy is expected to increase with hypervector dimensionality before plateauing, while latency and energy rise monotonically.  

\begin{figure}
    \centering
    \includegraphics[width=\textwidth]{ 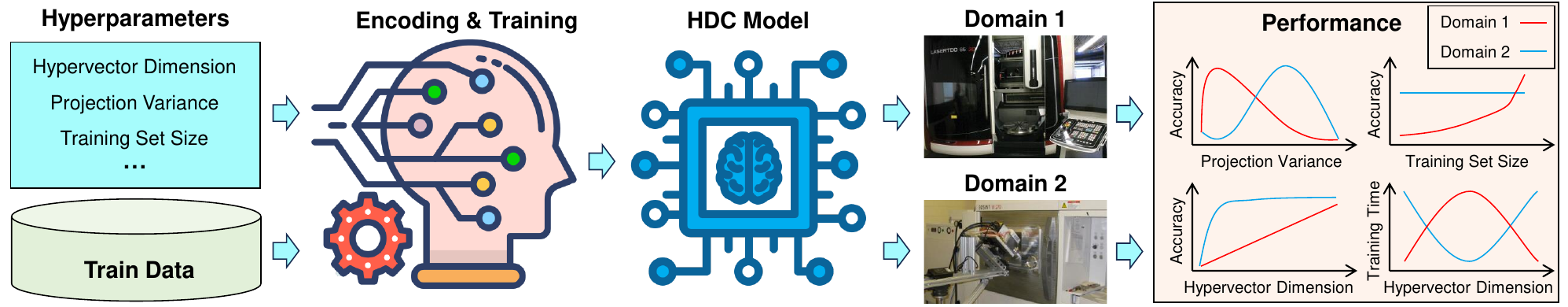}
    \caption{Domain-dependent HDC tuning. The optimal hyperparameters, such as hypervector dimensionality, projection variance, and training set size for maximizing accuracy and minimizing latency, vary across different application domains.}
    \label{fig:introduction}
\end{figure}

However, closer inspection reveals inconsistencies across domains, raising questions about the universality of these trade-offs(see Fig.~\ref{fig:introduction}). In this work, we demonstrate that this assumption does not always hold. Using two representative smart manufacturing problems, namely signal-based quality monitoring in Computer Numerical Control (CNC) machining and image-based defect detection in Laser Powder Bed Fusion (LPBF) additive manufacturing, we show that identical HDC parameters can yield qualitatively different performance profiles. This sensitivity is not limited to hypervector dimensionality but also extends to encoding choices such as projection variance or channel inclusion, activation and transformation functions such as linear projections versus nonlinear random Fourier features, and training dynamics such as the number of available samples and extent of retraining required. Some effects, like the dependence of inference latency on hypervector size, can be captured analytically, whereas others, like retraining cost, emerge from complex interactions among multiple factors and cannot be fully expressed in closed form. As a result, HDC behavior reflects a multi-input, multi-output optimization problem whose relationships between parameters and performance metrics are partly predictable and partly empirical.  

To explain these findings, we combine analytical and empirical perspectives. First, we present a formal complexity analysis of inference, training, and retraining, showing that operations scale linearly with feature size and dimensionality but interact with retraining frequency in non-monotonic ways. These dynamics cannot be captured with a simple closed-form expression, highlighting the need for empirical evaluation. Second, we validate these trade-offs experimentally on CNC and LPBF tasks, demonstrating that parameter-performance relationships differ qualitatively across domains. Together, these results show that there is no universal parameter setting and that domain-specific tuning is required to fully exploit HDC in resource-constrained edge environments.  

The main contributions of this paper are as follows:
\begin{enumerate}
\item We provide the first systematic analytical and empirical evidence challenging the widely held assumption of invariance in HDC parameter-performance landscapes. By comparing two representative manufacturing tasks, we show that the functional relationship between hyperparameters and performance metrics can differ qualitatively across domains.  
\item We conduct a formal complexity analysis of inference, training, and retraining in HDC, demonstrating that although operations scale linearly with feature size and dimensionality, the interplay between memorization capacity, retraining frequency, and computational cost produces non-monotonic trade-offs. This explains why optimal hyperparameters cannot be derived analytically and must be determined empirically.  
\item We empirically evaluate domain-specific tuning under multi-objective constraints of accuracy, inference latency, training time, and energy consumption. Results show that with proper domain-aware parameter selection, HDC outperforms deep neural networks and transformer baselines in latency and energy efficiency, underscoring its potential for real-time deployment in smart manufacturing.  
\end{enumerate}

The remainder of this paper is structured as follows. Section~\ref{sec:Research Background} reviews related work. Section~\ref{sec:Research Methodology} introduces the HDC framework and analytical complexity analysis. Section~\ref{sec:Experimental Design} details datasets and baselines. Section~\ref{sec:Experimental Results} reports findings, followed by discussion in Section~\ref{sec:Discussion and Limitations}. Section~\ref{sec:Conclusion and Future Works} concludes the paper and outlines future directions.

\section{Research Background}
\label{sec:Research Background}

The demand for real-time data processing has surged in recent years, driven by the growing need for instantaneous decision-making and operational efficiency in digital ecosystems~\cite{lavanya2024evolving}. Organizations increasingly depend on timely analytics to remain competitive and responsive to changing conditions~\cite{ranjan2021big}. The capability to analyze data in real-time is vital for uncovering actionable insights and enhancing system responsiveness, and deep learning has become a critical enabler of this capability~\cite{modupe2024reviewing}. However, as deep learning architectures grow larger to improve accuracy, their energy consumption scales up dramatically, and the operational cost and environmental impact of deploying such models at scale has become a critical concern.  

Edge computing has emerged as a direct response to these challenges by shifting computation and storage closer to the point of data generation~\cite{hamdan2020edge}. Unlike cloud-centric models that rely on centralized servers, edge computing deploys processing units near the network periphery, often embedded within industrial devices or gateways~\cite{simic2021towards}. This architectural decentralization reduces communication delays, lowers bandwidth demands, and enables responsiveness under strict latency constraints~\cite{shumba2022leveraging}. Yet, edge devices often operate under severe energy and resource limitations, where hundreds of inferences per second may need to be performed on hardware restricted to only tens of watts. These conditions motivate the development of algorithms that are not only accurate but also computationally and energetically frugal.

Traditional efforts to adapt ML to edge constraints have focused on compressing deep learning models. Approaches include pruning redundant connections~\cite{han2015learning}, quantizing weights to lower precision~\cite{wang2019haq}, tensor decomposition for compact representations~\cite{kim2015compression}, and knowledge distillation from large teacher models to lightweight students~\cite{tung2019similarity}. Neural architecture search has also been applied to automatically identify efficient structures~\cite{cai2018proxylessnas}. While effective in reducing inference cost, these approaches often sacrifice accuracy and still depend on backpropagation-based training, which is computationally expensive in resource-constrained industrial environments.  

HDC offers a fundamentally different alternative. Instead of relying on gradient descent, HDC represents data as high-dimensional hypervectors and processes them using simple operations such as binding, bundling, and permutation~\cite{aygun2023learning,masum2025ams}. Training requires only one-shot or few-shot updates, and inference reduces to vector similarity comparisons. These operations are parallelizable, hardware-friendly, and inherently energy-efficient~\cite{jalil2025privacy,desislavov2023trends}, making HDC an attractive candidate for smart manufacturing applications that demand both responsiveness and efficiency.  

The encoding stage is central to HDC performance. Raw input features must be mapped into hypervectors, and choices such as random projection , random Fourier features, quantization-based methods, or symbolic encodings significantly influence downstream performance~\cite{rahimi2007random,jalil2023hyperdimensional}. Hyperparameters including hypervector dimensionality, projection variance, activation functions, and retraining frequency directly shape accuracy, inference latency, training time, and energy consumption~\cite{piran2025privacy}. For instance, increasing dimensionality enriches representational capacity but raises inference cost, while lower dimensionality reduces latency but may lead to higher retraining overhead due to limited memorization capacity.  

Several optimization methods have been explored to manage these trade-offs. MicroHD~\cite{ponzina2024microhd} co-tunes multiple hyperparameters iteratively to maintain accuracy while reducing cost. MIMHD~\cite{kazemi2021mimhd} accelerates binding and bundling using in-memory operations with substantial energy savings. Other work has examined binarization for lightweight encoding~\cite{hernandez2021onlinehd}, exclusive versus inclusive encoding schemes~\cite{piran2025explainable,piran2024privacy}, and hardware- or privacy-aware extensions of HDC~\cite{piran2025privacy,chen2025federated}. While these studies confirm the importance of hyperparameter selection, they often optimize a single metric in isolation or assume that parameter-performance landscapes follow similar trends across domains.  

This assumption has not been systematically tested in smart manufacturing. Data in this domain span very different modalities, from temporal signals in CNC machining to high-resolution images in LPBF defect monitoring. If the parameter–performance relationships in HDC differ qualitatively between modalities, then heuristics tuned for one task may fail when applied to another. Addressing this gap requires systematic evaluation of HDC trade-offs under manufacturing-specific constraints, with a focus on multi-objective optimization that jointly balances accuracy, latency, training time, and energy efficiency.

\section{Research Methodology}
\label{sec:Research Methodology}

\subsection{Hyperdimensional Computing}
\label{sec:Hyperdimensional Computing}

 \begin{figure}
    \centering
    \includegraphics[width=\textwidth]{ 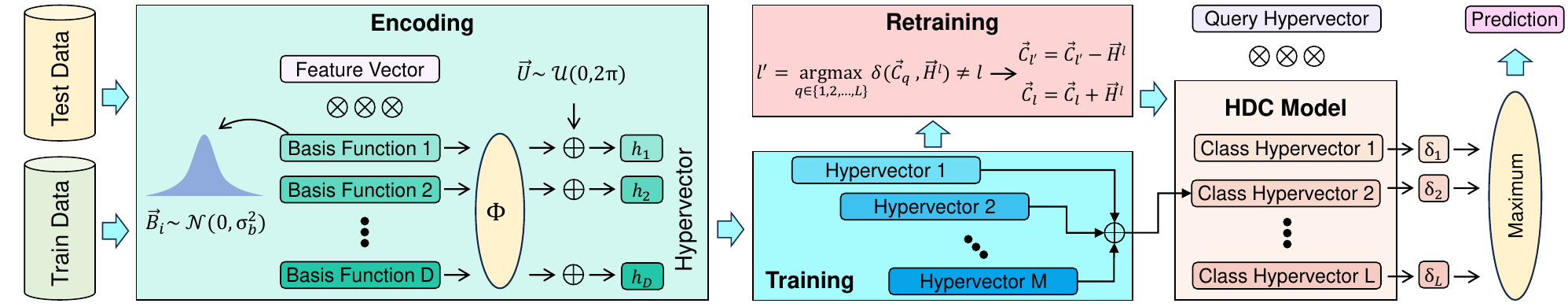}
    \caption{The HDC framework, illustrating the four core phases: encoding, training, inference, and retraining.}
    \label{fig:hd_framework}
\end{figure}

HDC encodes, stores, and processes information using vectors with thousands of dimensions, known as hypervectors. By exploiting distributed representations and simple algebraic operations, HDC provides a computational paradigm that is simultaneously lightweight and robust. The HDC pipeline is illustrated in Figure~\ref{fig:hd_framework} and consists of four stages, namely encoding, training, inference, and retraining.

In the \emph{encoding} stage, input feature vectors are transformed into high-dimensional hypervectors through randomized or structured projections. These hypervectors serve as the atomic building blocks for learning. During the \emph{training} stage, class prototypes are formed by aggregating the encoded hypervectors corresponding to each label. In \emph{inference}, unseen inputs are encoded and compared against the stored prototypes to identify the most similar class. Finally, \emph{retraining} incrementally updates class prototypes using misclassified samples, enabling continuous adaptation to evolving data.  

Each stage contributes differently to the computational profile of HDC. Encoding and inference costs scale with the input dimensionality and hypervector size, while training and retraining depend additionally on the number of samples and correction steps. These relationships are formally analyzed in Section~\ref{sec:Analytical Complexity}. This modular design allows HDC to flexibly adapt its accuracy–latency–energy trade-offs through the choice of encoding type, hypervector dimensionality, and retraining policy, making it particularly suitable for resource-constrained deployment in smart manufacturing environments.

\subsubsection{Encoding}
\label{sec:encoding}

During the encoding stage, input feature vectors are projected into a high-dimensional space to generate hypervectors that distribute information across dimensions. As illustrated in Figure~\ref{fig:hd_framework}, an input vector \(\vec{X} = \{x_{j}\}_{j=1}^{J}\) is transformed into a hypervector \(\vec{H} = \{h_i\}_{i=1}^D\) using a collection of randomly generated basis vectors \(\vec{B}_i\). Each basis vector is drawn from a Gaussian distribution with zero mean and standard deviation \({\sigma}_{b}\). To increase variability, a random offset vector \(\vec{U} = \{u_i\}_{i=1}^D\) is sampled uniformly from \(\mathcal{U}(0, 2\pi)\).  

The general form of the encoding function is expressed as
\begin{equation}
h_i = \phi(\vec{X} \cdot \vec{B}_i + u_i)
\label{eq:encoding_general}
\end{equation}
where \(\phi(\cdot)\) represents the activation function applied to generate the elements of the hypervector. In Random Projection (RP), a linear mapping is applied with \(\phi(x) = x\), which yields
\begin{equation}
h_i = \vec{X} \cdot \vec{B}_i 
\label{eq:encoding_rp}
\end{equation}

In Random Fourier Features (RFF), nonlinearity is introduced through a cosine transformation, resulting in
\begin{equation}
h_i = \cos(\vec{X} \cdot \vec{B}_i + u_i)
\label{eq:encoding_rff}
\end{equation}

These encoding approaches enable the transformation of real-valued features into high-dimensional representations while maintaining a balance between computational efficiency and representational expressiveness. Once generated, hypervectors are consistently used throughout training and inference, which makes HDC particularly well suited for applications such as defect detection and quality monitoring in smart manufacturing systems.

\subsubsection{Training}
\label{sec:training}

The training stage in HDC constructs a representative hypervector for each class, known as the class hypervector \(\vec{C}_l\), where \(l \in \{1, 2, \dots, L\}\) and \(L\) is the total number of classes. This prototype is obtained by aggregating all hypervectors \(\vec{H}^l_m\) that belong to class \(l\). The operation can be written as
\begin{equation}
\vec{C}_l = \sum_{m=1}^{M} \vec{H}^{l}_{m}
\label{eq:formClassHypervector}
\end{equation}
where \(M\) denotes the number of training samples in class \(l\), and each \(\vec{H}^l_m\) is a hypervector generated during the encoding stage.  

This aggregation, often described as superposition in the HDC literature, combines the statistical patterns of all members of a class into a single distributed representation. The resulting class hypervector provides a robust and noise-tolerant abstraction that generalizes across intra-class variations. At the same time, the training procedure relies only on additions in high-dimensional space, which keeps computational requirements low compared to gradient-based methods.  

The construction of class hypervectors establishes the memory structure that will later be used during inference and retraining. This makes training both efficient and directly compatible with edge-deployed applications, such as predictive maintenance and defect detection in smart manufacturing systems, where learning must be fast, incremental, and resource-conscious.

\subsubsection{Inference}
\label{sec:inference}

In the inference stage, the HDC model assigns a class label to a query input by comparing its hypervector \(\vec{H}\) with the set of class hypervectors \(\vec{C}_l\) created during training. The comparison is performed using cosine similarity, which measures the directional alignment between vectors in high-dimensional space. The similarity score is expressed as
\begin{equation}
\delta(\vec{C}_{l}, \vec{H}) = \frac{\vec{C}_{l} \cdot \vec{H}}{||\vec{C}_{l}|| \cdot ||\vec{H}||}.
\label{eq:inference similarity}
\end{equation}

The predicted class is the one whose prototype achieves the maximum similarity with the query hypervector, given by
\begin{equation}
\hat{l} = \mathop{\mathrm{argmax}}_{l \in \{1, 2, \dots, L\}} \delta(\vec{C}_{l}, \vec{H}).
\label{eq:hdc_prediction}
\end{equation}

HDC inference is computationally efficient because the norm of the query hypervector \(||\vec{H}||\) is constant across comparisons, and the norms of class hypervectors \(||\vec{C}_l||\) can be precomputed during training. As a result, the procedure reduces to evaluating dot products, which are lightweight, fast, and well suited to parallel hardware execution.  

This efficiency is particularly valuable for edge-based smart manufacturing applications, where real-time decision making is required under strict energy and latency constraints. By relying on simple vector operations, HDC supports accurate and low-latency classification directly on resource-constrained devices, eliminating the need for cloud offloading while enabling timely responses for tasks such as anomaly detection and quality monitoring.

\subsubsection{Retraining}
\label{sec:retraining}

Retraining improves the adaptability and robustness of HDC models by incrementally correcting errors that remain after initial training. The process begins by comparing the encoded hypervectors of training samples with the stored class prototypes. If a sample from class \(l\) is incorrectly classified as class \(l'\), the misprediction can be expressed as
\begin{equation}
l' = \mathop{\mathrm{argmax}}_{q \in \{1, 2, \dots, L\}} \delta(\vec{C}_{q}, \vec{H}^{l}),
\label{eq:mis_prediction}
\end{equation}
where \(\vec{H}^l\) denotes the hypervector of the misclassified instance.  

The retraining step then updates the prototypes by subtracting the contribution of \(\vec{H}^l\) from the incorrect class and adding it to the correct class:
\begin{equation}
\begin{aligned}
\vec{C}_{l'} &= \vec{C}_{l'} - \vec{H}^{l}, \\
\vec{C}_{l}  &= \vec{C}_{l} + \vec{H}^{l}.
\end{aligned}
\label{eq:updateclasshv}
\end{equation}

Through this adjustment, class hypervectors become progressively more aligned with the true data distribution, reducing the likelihood of repeated misclassifications. Retraining is usually repeated over several epochs, gradually refining decision boundaries without requiring a complete model rebuild.  

This mechanism is particularly valuable in smart manufacturing environments, where sensor data can drift over time due to tool wear, material variation, or changing operating conditions. By supporting lightweight and incremental updates, HDC retraining enables continuous adaptation on edge devices, ensuring high accuracy while maintaining low computational and energy overhead.

\subsection{Problem Statement}
\label{sec:Problem Statement}

In the context of smart manufacturing, the encoding stage of HDC plays a decisive role in shaping overall performance. Applications such as defect detection in LPBF and anomaly monitoring in CNC machining demand hundreds of inferences per second under strict energy and latency budgets. Meeting these requirements depends on how raw sensor features are projected into the high-dimensional hyperspace. A central design choice in this process is the transformation function, typically selected between RP and RFF. RP applies a linear mapping, preserving the structure of the input space, while RFF introduces a nonlinear cosine activation to capture richer feature interactions. The choice between these functions directly influences representational power, generalization capability, and computational cost, and must therefore be tuned in relation to both task complexity and the hardware constraints of industrial edge devices.

Beyond the transformation type, the dispersion of the random basis vectors, controlled by the standard deviation parameter \(\sigma_b\), strongly affects the discriminability of the encoded space. Larger values of \(\sigma_b\) produce more distinct hypervectors, yielding exclusive encodings that improve class separability, which is beneficial for fine-grained quality monitoring. Smaller values of \(\sigma_b\) create inclusive encodings, generating greater overlap between hypervectors and favoring generalization across related conditions, such as variations in tool vibration or material microstructure. Thus, tuning \(\sigma_b\) provides a mechanism to balance specificity against generalizability in manufacturing tasks where both accuracy and adaptability are essential.

The dimensionality of the hypervectors, denoted by \(D\), introduces another crucial trade-off. Higher values of \(D\) expand representational capacity and can increase classification accuracy, but this comes at the cost of greater inference latency, longer training times, and higher energy consumption since both encoding and similarity computations scale linearly with \(D\). Formally, similarity operations in inference incur a complexity proportional to \(\mathcal{O}(LD)\), where \(L\) is the number of classes, while encoding scales as \(\mathcal{O}(JD)\) with \(J\) input features. Conversely, smaller values of \(D\) reduce latency and energy consumption but may limit discriminability, leading to more frequent misclassifications and greater retraining overhead. 

In practice, these parameters including transformation type, projection variance \(\sigma_b\), and hypervector dimensionality \(D\) must be tuned jointly. A large \(D\) combined with an exclusive encoding can reduce the need for retraining by improving initial classification accuracy, but the resulting computational cost may exceed the energy limits of edge devices in industrial environments. In contrast, compact encodings reduce latency but risk unstable performance over time. This setting defines a multi-objective optimization problem in which accuracy, inference speed, training efficiency, and energy usage are closely interdependent. The challenge is to identify configurations that are tailored to the domain and achieve the best balance, ensuring reliable real-time performance in smart manufacturing applications.

\subsection{Analytical Complexity}
\label{sec:Analytical Complexity}

This section analyzes the computational complexity of the three principal stages of the HDC pipeline, namely inference, training, and retraining. Let \( J \) denote the dimensionality of the input feature vectors, \( D \) the dimensionality of the hypervectors, \( L \) the number of classes, \( N \) the number of training samples, and \( P \) the total number of correction steps performed during retraining.

\subsubsection{Inference}

Inference consists of two main steps. The first is encoding the query input into a \( D \)-dimensional hypervector, which requires multiplying the input vector of size \( J \) with a projection basis of size \( D \). This yields a computational cost of
\begin{equation}
\mathcal{O}(JD).
\end{equation}
If a nonlinear activation such as RFF encoding is applied, this adds an extra cost of \(\mathcal{O}(D)\).  

The second step is similarity computation between the encoded query hypervector and each of the \( L \) class hypervectors. Each comparison involves a dot product of dimension \( D \), resulting in
\begin{equation}
\mathcal{O}(LD).
\end{equation}
Therefore, the overall inference complexity is
\begin{equation}
\mathcal{O}(JD + LD).
\label{eq:hdc_inference_complexity}
\end{equation}

\subsubsection{Training}

In training, each of the \( N \) samples is encoded into a \( D \)-dimensional hypervector, leading to
\begin{equation}
\mathcal{O}(NJD).
\end{equation}
Once encoded, the hypervectors are aggregated by vector addition to form class prototypes. Since each addition requires \(\mathcal{O}(D)\) operations, the aggregation cost for all samples is
\begin{equation}
\mathcal{O}(ND).
\end{equation}
Thus, the total training complexity is
\begin{equation}
\mathcal{O}(NJD + ND) = \mathcal{O}(ND(J+1)).
\label{eq:hdc_training_complexity}
\end{equation}

\subsubsection{Retraining}

Retraining updates class prototypes whenever misclassifications occur. For each of the \( P \) correction steps, the procedure involves re-encoding the sample with cost \(\mathcal{O}(JD)\), computing similarity with all class hypervectors with cost \(\mathcal{O}(LD)\), and updating the relevant prototypes with cost \(\mathcal{O}(D)\). Hence the overall retraining complexity is
\begin{equation}
\mathcal{O}(P(JD + LD + D)) = \mathcal{O}(PD(J + L + 1)).
\label{eq:hdc_retraining_complexity}
\end{equation}

These results show that inference, training, and retraining all scale linearly with hypervector dimensionality \( D \). This linear scaling highlights why HDC is attractive for deployment in edge-based smart manufacturing, where energy and latency budgets are highly constrained yet fast adaptation to new data is required.

\subsection{Optimization Framework}
\label{sec:optimization}

The objective of this work is to optimize the configuration of the HDC model in order to achieve high classification accuracy while simultaneously minimizing inference time, training time, and energy consumption. To formalize this, we adopt a multi-objective optimization formulation, where the goal is to jointly maximize accuracy and minimize latency and resource usage. The optimization problem is defined as follows:

\begin{equation}
\begin{aligned}
& \underset{\theta = (t,\,D,\,\sigma_b)}{\text{maximize}} 
& & \Big[f_{\mathrm{acc}}(\theta),\; -T_{\mathrm{inference}}(\theta),\; -T_{\mathrm{train}}(\theta),\; -E_{\mathrm{train}}(\theta)\Big] \\
& \text{subject to}
& & f_{\mathrm{acc}}(\theta) \ge f_{\min}, \\
& 
& & T_{\mathrm{inference}}(\theta) \le I_{\max}, \\
& 
& & T_{\mathrm{train}}(\theta) \le T_{\max}, \\
& 
& & E_{\mathrm{train}}(\theta) \le E_{\max}, \\
& 
& & t \in \{\mathrm{RP},\, \mathrm{RFF}\},\quad D \in \mathbb{N},\quad \sigma_b \in \mathbb{R}^+.
\end{aligned}
\label{eq:opt_params}
\end{equation}

\noindent where the parameters are defined as follows:

\begin{itemize}
    \item \(t\): encoding type, chosen between Random Projection (RP) and Random Fourier Features (RFF);  
    \item \(D\): hypervector dimensionality, which determines representational capacity;  
    \item \(\sigma_b\): standard deviation of the Gaussian basis vectors, controlling the spread of projections and influencing separability of the encoded space;  
    \item \(f_{\mathrm{acc}}(\theta)\): classification accuracy obtained under configuration \(\theta\);  
    \item \(T_{\mathrm{inference}}(\theta)\): average inference latency;  
    \item \(T_{\mathrm{train}}(\theta)\): training time required to construct class hypervectors;  
    \item \(E_{\mathrm{train}}(\theta)\): energy consumption measured during training;  
    \item \(f_{\min}\): minimum acceptable classification accuracy;  
    \item \(I_{\max}\): maximum allowable inference latency;  
    \item \(T_{\max}\): maximum allowable training time;  
    \item \(E_{\max}\): maximum allowable training energy consumption.  
\end{itemize}

Each candidate configuration \(\theta = (t, D, \sigma_b)\) is evaluated through a complete training and testing pipeline that includes encoding of input features, construction of class hypervectors, and measurement of accuracy, inference latency, training time, and energy consumption. To efficiently explore this space, we employ Bayesian optimization, which is well-suited for non-convex, mixed discrete–continuous problems. At each iteration, a candidate configuration is selected based on an acquisition function, evaluated on the full pipeline, and its performance metrics are recorded. These results are then used to update the surrogate model, guiding subsequent iterations toward configurations that achieve the best balance across objectives.

This framework ensures that optimization is not biased toward a single metric but instead reflects the true multi-objective nature of HDC deployment in smart manufacturing. By explicitly considering the coupling between accuracy, latency, training cost, and energy usage, the method identifies domain-aware parameter settings that meet industrial constraints for real-time monitoring and control.

\section{Experimental Design}
\label{sec:Experimental Design}

We evaluate the proposed framework using two real-world case studies from smart manufacturing, each formulated as a supervised classification task. The first case study addresses defect detection in LPBF, an image-based problem that leverages high-speed camera data to identify process anomalies. The second case study focuses on quality monitoring in CNC machining, which relies on time series sensor signals to detect deviations in operational quality. These complementary tasks allow us to examine the behavior of HDC across distinct data modalities, namely images and signals, while also highlighting the importance of domain-aware parameter optimization. The following subsections describe the experimental setup and dataset characteristics for each task.

\subsection{Laser Powder Bed Fusion Task}
\label{sec:lpbf task}

\begin{figure}
    \centering
    \includegraphics[width=\textwidth]{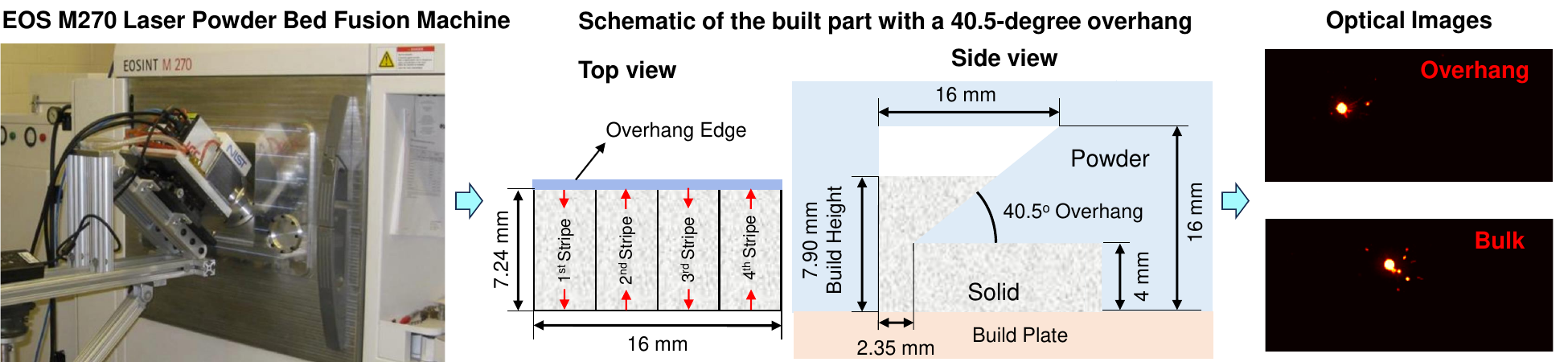}
    \caption{The experimental setup, showing the EOS M270 LPBF machine~\cite{lane2016thermographic}, the top view schematic of the built part, and the side view schematic with dimensions, including a 40.5 degree overhang. Moreover, captured optical images from the process of the overhang and the bulk areas.}
    \label{fig:lpbf_dataset}
\end{figure}

This task uses experimental data collected from an EOS M270 LPBF machine located at the National Institute of Standards and Technology (NIST), as shown in Figure~\ref{fig:lpbf_dataset}. A high-speed visible-range camera was integrated into the build chamber to support real-time process monitoring. The camera was positioned in the upper right region of the chamber and mounted using a custom sealed through-port that protected it from the build environment. Although no precise calibration of the camera’s position or focal distance was performed, its configuration enabled comprehensive observation of melt pool dynamics and spatter formation.

The camera is equipped with a silicon-based sensor array that provides a resolution of 1.2 megapixels and a spectral sensitivity ranging from 300 to 950 nm. It operates at frame rates of up to 400 frames per second. The captured images were cropped to a fixed size of \(256 \times 256\) pixels and sampled at 1000 frames per second. These recordings capture essential spatial and temporal features of the build process, including both bulk regions and overhang structures. The experimental configuration reflects typical industrial LPBF workflows and is employed here to enable supervised learning for image-based defect classification~\cite{lane2016thermographic}.

The example part in this study was fabricated using process parameters representative of industrial LPBF. The settings included a hatch spacing of 0.1 mm, a stripe width of 4 mm with a 0.1 mm overlap, and a layer thickness of 20~\(\mu\)m. The laser operated at a scan speed of 800 mm/s with an infill power of 195 W. Contour scans began at the part corners and proceeded counterclockwise along the perimeter, with distinct laser powers assigned to each pass: 100 W for pre-contour and 120 W for post-contour scans, both executed at 800 mm/s.

The dataset was constructed by monitoring 45 build layers of a 16 mm tall nickel alloy 625 part. Particular attention was given to distinguishing overhang and bulk regions within each layer. Overhang regions were defined as the last two scan vectors adjacent to edge formation, excluding contour scans, while the remaining regions were categorized as bulk. A stripe-based scanning strategy was applied, with the orientation alternating by 90° between consecutive layers. Once the build height exceeded 4 mm, the overhang region was subjected to four scan passes per layer.

Sensor analysis concentrated on three representative build heights of 6.06 mm, 7.90 mm, and 9.70 mm. These stages captured the structural evolution of the unsupported 40.5° overhang, a common challenge in LPBF caused by thermal distortion and the absence of support material. The powder used was vendor-supplied nickel alloy 625 with an average particle diameter of 37.8~\(\mu\)m, deposited on a custom alloy 625 substrate. For the experiments presented in this paper, the focus was placed on a single build layer at 7.90 mm, which provided insight into the thermal and geometric features associated with overhang formation. A schematic side view of the printed structure is presented in Figure~\ref{fig:lpbf_dataset}.

To enable in-situ monitoring, the LPBF process was observed using a high-speed visible-spectrum camera capable of high-resolution and rapid-frame capture. The camera, mounted inside the build chamber, continuously recorded thermal activity during layer-by-layer melting of the powder. This imaging setup facilitated detailed observation of melt pool dynamics, spatter, and heat distribution. All frames were cropped to a fixed resolution during acquisition to ensure consistency across the dataset.

The collected images were then organized into a classification scheme consisting of eight distinct classes. These classes corresponded to specific combinations of overhang and bulk regions distributed across four stripe segments. The class structure was designed to capture meaningful thermal variations observed during the build, enabling practical relevance for detecting defects that could compromise the mechanical integrity or surface quality of the printed component. Each class thus reflected a unique thermal profile, allowing fine-grained analysis of localized anomalies and supporting targeted defect detection in industrial LPBF applications.

\subsection{Computer Numerical Control Task}
\label{sec:cnc task}

Geometric precision in manufactured components is fundamental for ensuring reliable mechanical performance, as it directly affects stiffness, strength, and resistance to failure~\cite{ramirez2023review}. Even small deviations from the intended dimensions can result in structural defects that compromise part functionality~\cite{naranjo2019tensile}. Continuous monitoring and control of dimensional accuracy during machining is therefore essential to maintaining product quality and reducing the risk of premature failure~\cite{fu2022machine}.

\begin{figure}
    \centering
    \includegraphics[width=\textwidth]{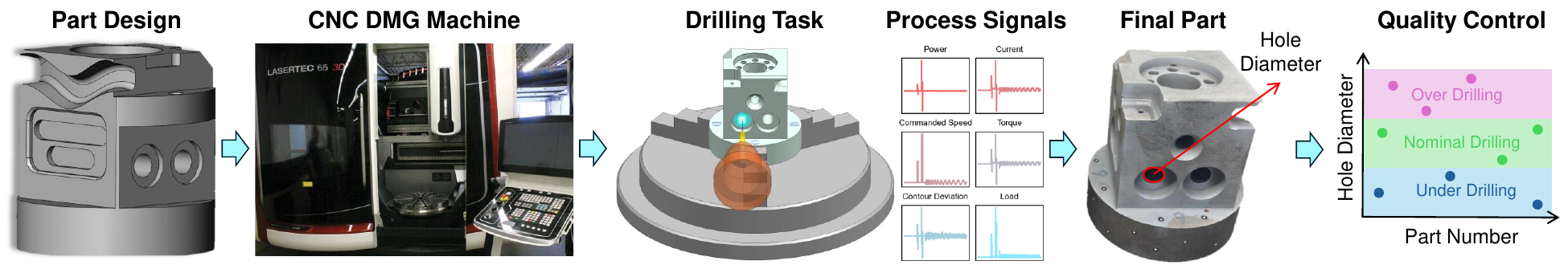}
    \caption{Experimental setup for the CNC drilling task, showing the steel workpieces, drilling locations, and representative signal channels collected during machining. Z-score-based labels (\textit{Under Drilling}, \textit{Nominal Drilling}, \textit{Over Drilling}) are assigned based on measured hole diameters.}

    \label{fig:cnc_dataset}
\end{figure}

In this study, drilling operations were performed on 18 steel blocks (1040 alloy) using a LASERTEC 65 DED hybrid CNC machine. Each block measured 76.2~mm $\times$ 76.2~mm $\times$ 76.2~mm, with a single hole drilled into each part, as illustrated in Figure~\ref{fig:cnc_dataset}. Post-process measurements of the hole diameters were used to compute Z-scores, which enabled classification of each part into one of three quality categories: \textit{Nominal Drilling} ($-1 \leq Z \leq 1$), \textit{Under Drilling} ($Z < -1$), and \textit{Over Drilling} ($Z > 1$).

During machining, real-time process signals were collected using a Siemens Simatic IPC227E data acquisition unit connected to the CNC machine. A total of 15 signals were captured, including load, torque, current, commanded speed, control differential, power, and encoder positions. These were sampled at 500~Hz across the spindle and five axes, yielding 90 signal channels per part.

To prepare the dataset for analysis, each part’s 90-channel signal stream was segmented using a sliding window of length 10. This preprocessing step generated fixed-length multivariate samples, where each sample contained 90 features over 10 time steps. The samples inherited the quality label of the corresponding part, thereby enabling detection of subtle geometric deviations in real-time CNC operations.

\subsection{Optimization Setup}
\label{sec:Optimization Setup}

To identify optimal HDC configurations for deployment in resource-constrained environments, we conducted a Bayesian optimization for both case studies. The objective was to maximize classification accuracy while minimizing inference latency, training time, and energy consumption.

The optimization explored combinations of the encoding type \( t \in \{\text{RP}, \text{RFF}\} \), hypervector dimensionality \( D \in [100, 50000] \), and standard deviation \( \sigma_b \in [0.01, 2.0] \) used in the Gaussian projection basis. For each candidate configuration, the HDC model was trained using perceptron-style updates for 20 epochs.

Energy consumption and training duration were measured during the optimization process, while inference time was obtained by encoding and classifying the entire test set. Bayesian optimization was executed for 50 episodes per task, with performance metrics including accuracy, training time, inference time, and energy consumption recorded for analysis. This setup enabled the identification of efficient domain-aware configurations that balance accuracy and efficiency for edge-based smart manufacturing systems.

\subsection{Evaluation Metrics}
\label{sec:evaluation_metrics}

To evaluate the performance of each HDC configuration during the optimization process, we monitored four primary metrics. Accuracy was measured as the proportion of correctly classified test samples, serving as the main indicator of predictive performance. Training time was defined as the total time required to encode the training set, initialize the class hypervectors, and perform all retraining epochs, reflecting the practicality of deployment under time-sensitive constraints. Inference time was recorded as the duration required to encode and classify the entire test set, and provided a direct measure of responsiveness for real-time operation on edge devices. Energy consumption was tracked during training to capture the overall computational efficiency of the configuration. These metrics jointly guided the multi-objective optimization process by quantifying the trade-offs between model accuracy, latency, training efficiency, and energy usage.  

\subsection{Hardware Environment}
\label{sec:hardware_software}

All experiments were conducted on a system equipped with an NVIDIA A100 GPU (40\,GB memory, 6912 CUDA cores, and 1.6\,TB/s memory bandwidth) and SSD storage. While this hardware provides ample throughput, the optimization process was explicitly designed to enforce latency and energy limits that mirror the requirements of edge deployment in smart manufacturing. Energy consumption and timing were measured during both training and inference to ensure that the reported results reflect conditions relevant to resource-constrained environments.

\begin{table}
\centering
\caption{Summary of CNC Task Benchmark Model Configurations}
\label{tab:cnc_benchmarks}
\renewcommand{\arraystretch}{1.2}

\begin{tabular}{l p{13.5cm}}  
\hline
\textbf{Model} & \textbf{Architecture Details} \\
\hline
CNN & Four convolutional blocks (Conv2d $\rightarrow$ BatchNorm $\rightarrow$ ReLU $\rightarrow$ MaxPool $\rightarrow$ Dropout), followed by two fully connected layers. Input shape: (1, 90, 10). Trained for 500 epochs using SGD with a learning rate of 1e$^{-3}$ and a batch size of 16. \\
ConvLSTM & Two 1D convolutional layers followed by a four-layer bidirectional LSTM (hidden dimension = 128) and a final linear classifier. Input shape: (1, 90, 10). Trained using the Adam optimizer. \\
RNN (LSTM) & Three-layer LSTM (hidden dimension = 128), followed by a fully connected classifier. Input shape: (90, 10). Trained using the Adam optimizer. \\
Transformer & One encoder block with four attention heads ($d_\text{model} = 64$), feed-forward dimension = 128, ReLU activation, and fixed sinusoidal positional encoding. Global mean pooling followed by a linear classifier. Trained using the Adam optimizer. \\
\hline
\end{tabular}
\end{table}

\subsection{Benchmark Models}
\label{sec:benchmark_models}

To benchmark the optimized HDC framework, we compared its performance with several widely used deep learning architectures tailored to the characteristics of each task. For the CNC task, which involves multivariate time-series signals, we implemented four sequence models. These included a Convolutional Neural Network (CNN), a Recurrent Neural Network (RNN), a Convolutional Long Short-Term Memory Network (ConvLSTM), and a Transformer-based model. Their configurations are summarized in Table~\ref{tab:cnc_benchmarks}, and they were evaluated in terms of accuracy, training time, inference latency, and energy consumption.  

For the LPBF task, which involves image-based defect detection, we selected four established computer vision models. These included AlexNet~\cite{krizhevsky2012imagenet}, ResNet50~\cite{he2016deep}, GoogLeNet (Inception-v1)~\cite{szegedy2015going}, and a Vision Transformer (ViT)~\cite{dosovitskiy2020image} with patch size 16 pretrained on ImageNet. Each model was fine-tuned using the same image dimensions and dataset splits as the HDC experiments, ensuring consistent comparisons across approaches.

\section{Experimental Results}
\label{sec:Experimental Results}

\subsection{Sensitivity Analysis}
\label{sec:Sensitivity Analysis}

\begin{figure}
\centering

\subfloat[]{\includegraphics[width=0.3\textwidth]{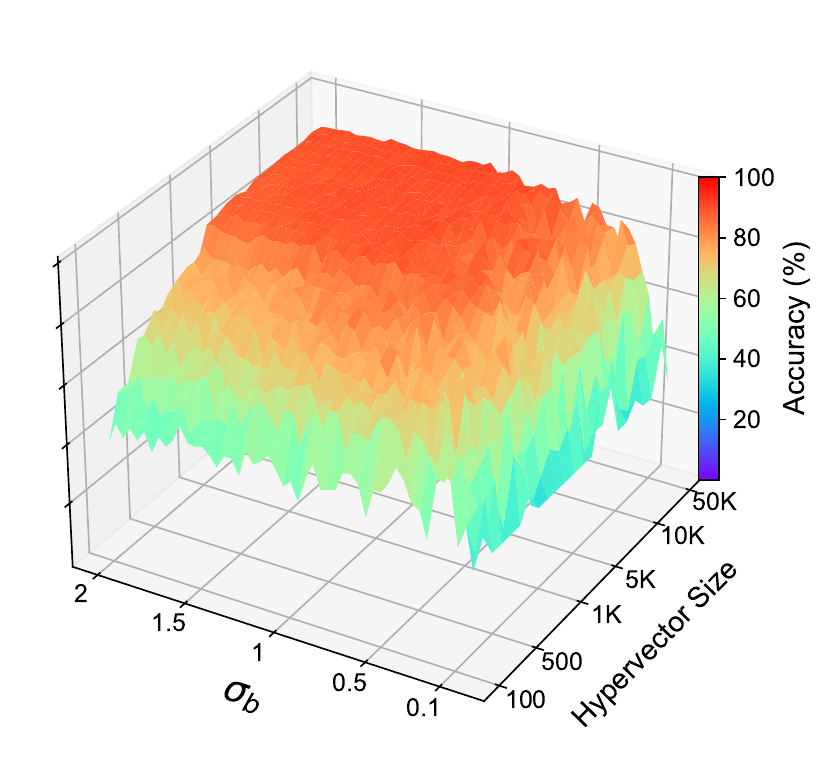}%
\label{fig:acc_3d_sub1}} 
\subfloat[]{\includegraphics[width=0.3\textwidth]{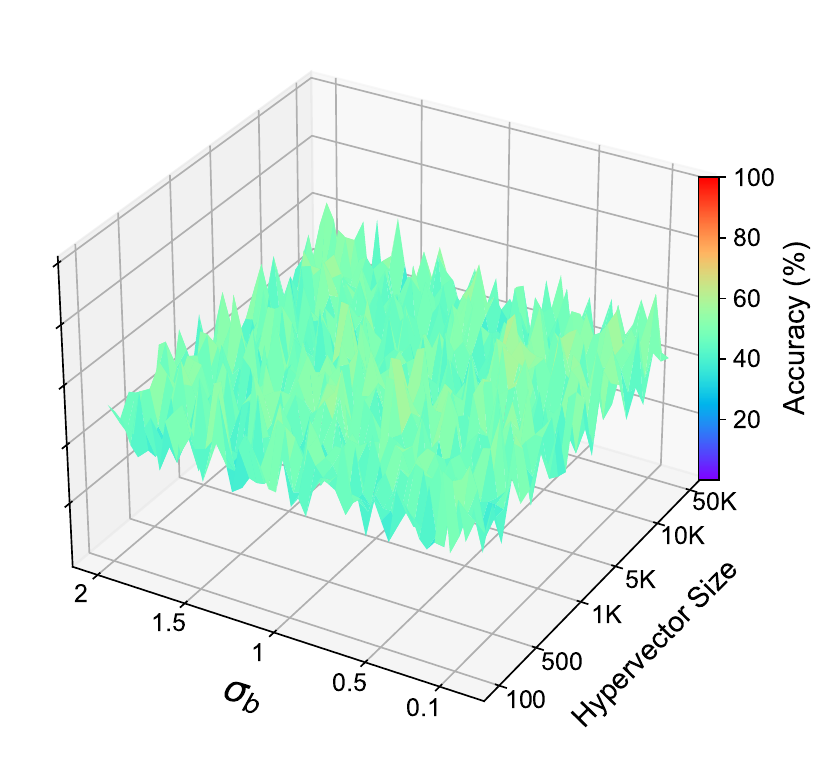}%
\label{fig:acc_3d_sub2}}

\subfloat[]{\includegraphics[width=0.3\textwidth]{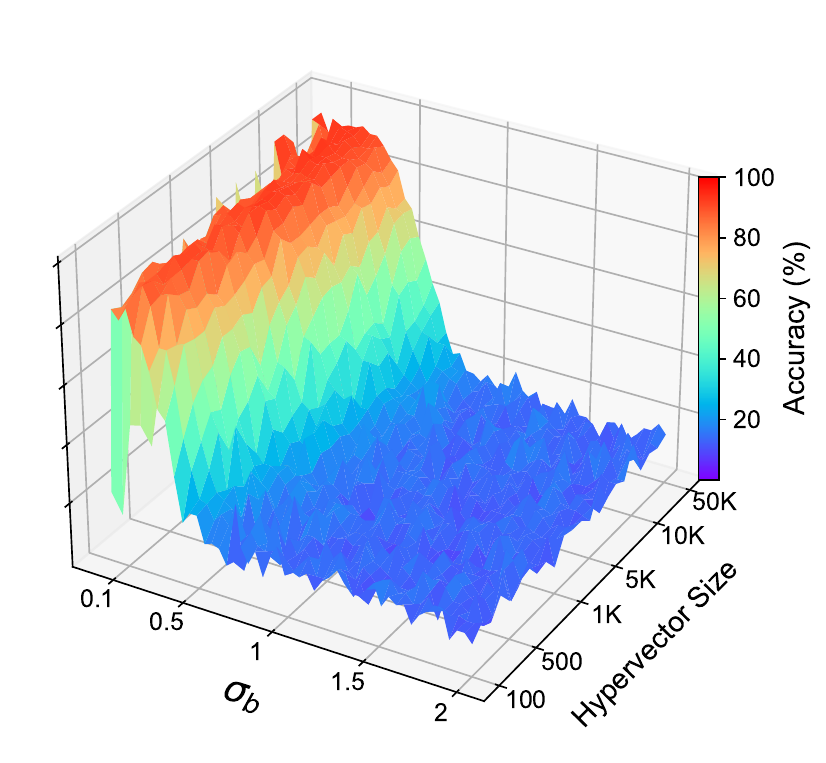}%
\label{fig:acc_3d_sub3}} 
\subfloat[]{\includegraphics[width=0.3\textwidth]{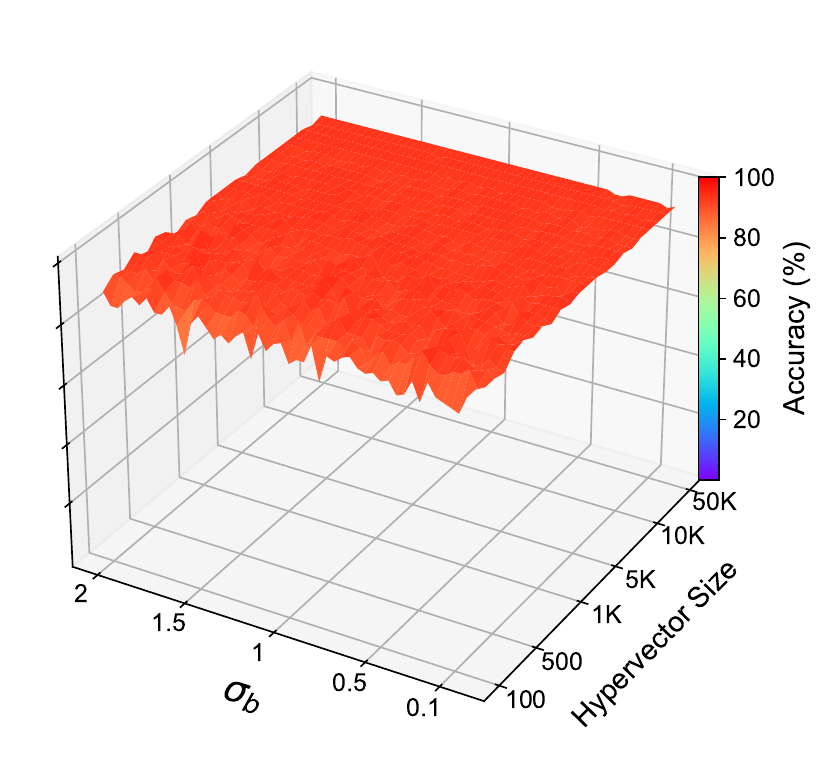}%
\label{fig:acc_3d_sub4}}

\caption{Classification accuracy for the CNC task using (a) Cosine and (b) RP encoding, and for the LPBF task using (c) Cosine and (d) RP encoding.}
\label{fig:acc_sig_d_3d}
\end{figure}

Figure~\ref{fig:acc_3d_sub1} illustrates the classification accuracy of the CNC task under RFF encoding across combinations of hypervector dimensionality \(D\) and projection standard deviation \(\sigma_b\). When \(\sigma_b = 0.1\), accuracy remains below \(75\%\) regardless of the hypervector size, indicating that inclusive encoding fails to capture sufficient discriminative information. In contrast, increasing \(\sigma_b\) to 2 substantially improves accuracy, reaching \(89.1\%\) at \(D = 10{,}000\). This result shows that exclusive encoding, obtained with a larger \(\sigma_b\), is more effective for signal-based CNC classification.  

As the hypervector dimensionality increases, accuracy improves because the larger hyperspace can encode more complex signal patterns. For example, with \(\sigma_b = 2\), accuracy rises from \(62.60\%\) at \(D = 500\) to \(74.86\%\) at \(D = 1{,}000\), then to \(87.83\%\) at \(D = 5{,}000\), and reaches \(89.10\%\) at both \(D = 10{,}000\) and \(D = 50{,}000\). The absence of further gains beyond \(D = 10{,}000\) indicates diminishing returns, while inference latency continues to grow. Thus, the practical choice is to select the smallest \(D\) that achieves the target accuracy, ensuring efficiency without unnecessary computational cost.

In contrast, Figure~\ref{fig:acc_3d_sub2} shows that RP encoding consistently produces lower accuracy, never surpassing \(67.70\%\) across all tested configurations. This outcome indicates that the CNC task, which relies on multivariate time-series signals, benefits from the nonlinearity introduced by RFF. The purely linear mapping of RP does not capture the discriminative patterns needed for reliable classification in this domain.

In contrast to the CNC task, where exclusive encoding is more effective, the LPBF task shows a clear preference for inclusive encoding. As illustrated in Figure~\ref{fig:acc_3d_sub3}, the accuracy of RFF encoding remains below \(23\%\) when \(\sigma_b = 2\), demonstrating that exclusive encoding fails to capture meaningful image features. By narrowing the basis spread to \(\sigma_b = 0.1\), accuracy increases substantially, reaching \(91\%\) at a hypervector size of \(D = 50{,}000\). These results suggest that inclusive encoding is better suited for representing the spatial patterns needed to classify LPBF images effectively.

Figure~\ref{fig:acc_3d_sub4} presents the accuracy of RP encoding for the LPBF task. Unlike the CNC case, where RP accuracy never exceeds \(68\%\), RP performs remarkably well on the LPBF dataset, achieving more than \(91\%\) accuracy with \(D = 2{,}000\) under both inclusive and exclusive settings. This makes RP the preferred encoding method for the image-based task, since it delivers high accuracy with relatively small hypervector dimensionality, thereby reducing inference time. In addition, RP avoids dependence on the parameter \(\sigma_b\), simplifying the optimization process and lowering uncertainty in model configuration.

\begin{figure}
\centering

\subfloat[]{\includegraphics[width=0.32\textwidth]{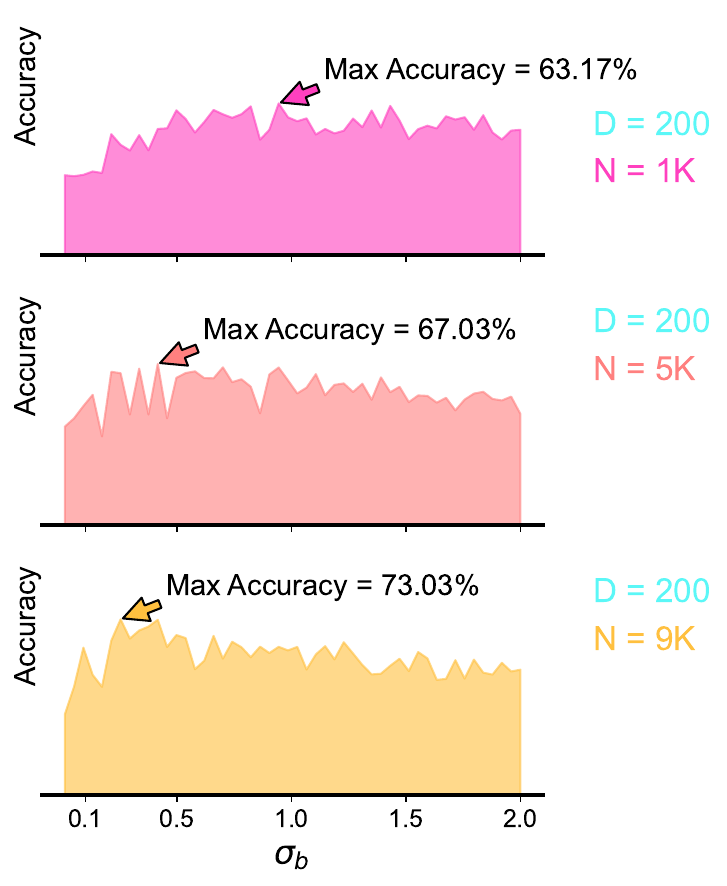}%
\label{fig:acc_starb_cnc_sub1}} \hfill
\subfloat[]{\includegraphics[width=0.32\textwidth]{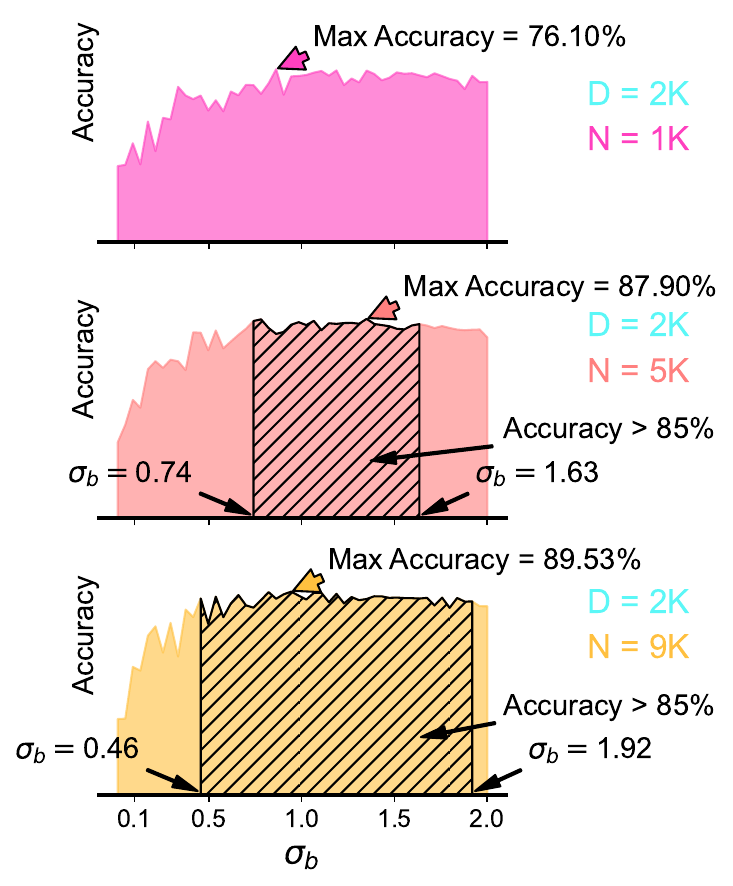}%
\label{fig:acc_starb_cnc_sub2}} \hfill
\subfloat[]{\includegraphics[width=0.32\textwidth]{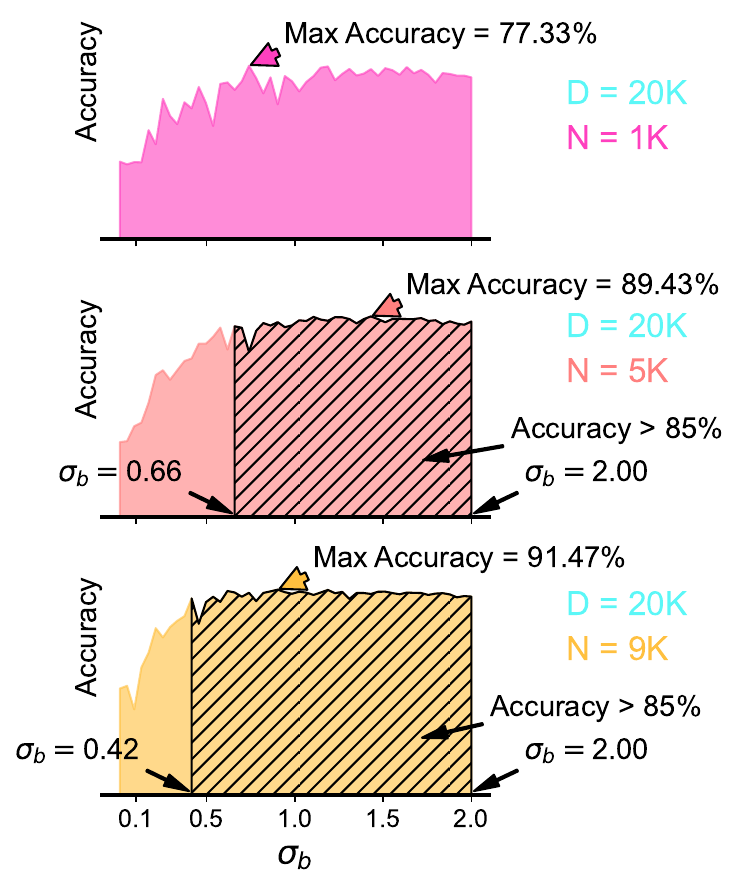}%
\label{fig:acc_starb_cnc_sub3}}

\caption{Effect of hypervector dimensionality and standard deviation of RFF basis functions on model accuracy for the CNC task, evaluated with (a) 200, (b) 2K, and (c) 20K training samples.}
\label{fig:acc_N_starb_cnc}
\end{figure}

Figure~\ref{fig:acc_N_starb_cnc} presents the effect of training set size on classification accuracy for the CNC task using RFF encoding. In Figure~\ref{fig:acc_starb_cnc_sub1}, where the hypervector dimensionality is fixed at a small value \( D = 200 \), the number of training samples \( N \) increases from 1K to 9K across different \(\sigma_b\) values. As expected, enlarging the training set improves accuracy, rising from \(63.17\%\) to \(73.03\%\), since additional data refines the class hypervectors and enhances generalization to unseen queries. Nonetheless, the limited hyperspace capacity at \(D = 200\) restricts the model’s ability to fully capture discriminative patterns, and performance saturates below \(75\%\).  

When the hypervector dimensionality is increased to \( D = 2{,}000 \), as shown in Figure~\ref{fig:acc_starb_cnc_sub2}, the model achieves substantial accuracy improvements. The maximum observed accuracy across all \(\sigma_b\) values rises to \(76.10\%\), \(87.90\%\), and \(89.53\%\) for training sizes \(N = 1\text{K}\), \(5\text{K}\), and \(9\text{K}\), respectively. Compared to the \(D = 200\) case, these correspond to absolute gains of \(12.93\%\), \(20.87\%\), and \(16.50\%\). This demonstrates that enlarging the hyperspace provides sufficient representational capacity to capture complex signal variations, allowing accuracy to scale more effectively with increasing training data.

Further expanding the hyperspace to \( D = 20{,}000 \), as illustrated in Figure~\ref{fig:acc_starb_cnc_sub3}, leads to additional gains in accuracy, reaching \(77.33\%\), \(89.43\%\), and \(91.47\%\) for training sizes \(N = 1\text{K}\), \(5\text{K}\), and \(9\text{K}\), respectively. While these results confirm the benefit of larger dimensionality in capturing complex signal dynamics, the improvements are accompanied by a marked increase in inference latency. This underscores the trade-off between representational capacity and computational efficiency, where optimal performance requires selecting a dimensionality that balances accuracy with practical constraints on latency and energy in smart manufacturing applications.  

The number of training samples \( N \) and the hypervector dimensionality \( D \) jointly determine how sensitive the system is to the choice of \(\sigma_b\). This sensitivity directly affects the robustness of model performance under different encoding configurations. Identifying an optimal \(\sigma_b\) that simultaneously meets accuracy and latency requirements is therefore nontrivial, particularly in scenarios where the model exhibits strong dependence on encoding parameters.  

For instance, if we set \(85\%\) as the minimum acceptable accuracy, increasing the hypervector dimensionality \(D\) broadens the interval of \(\sigma_b\) values that satisfy this criterion. When \( N = 5\text{K} \) and \( D = 2{,}000 \), the system maintains at least \(85\%\) accuracy for \(\sigma_b \in [0.74,\, 1.63]\). Expanding \( D \) to \( 20{,}000 \) widens the valid interval to \([0.66,\, 2.00]\), thereby providing greater flexibility in parameter selection without sacrificing predictive performance.  

A similar trend is observed when increasing the number of training samples \(N\). For a fixed hypervector dimensionality of \(D = 2{,}000\), expanding the training set from \(5\text{K}\) to \(9\text{K}\) enlarges the valid range of \(\sigma_b\) from \([0.74,\, 1.63]\) to \([0.46,\, 1.92]\). This result indicates that larger datasets enhance the robustness of the system to encoding parameter variations and reduce the uncertainty associated with tuning \(\sigma_b\) during optimization.

\begin{figure}
\centering

\subfloat[]{\includegraphics[width=0.32\textwidth]{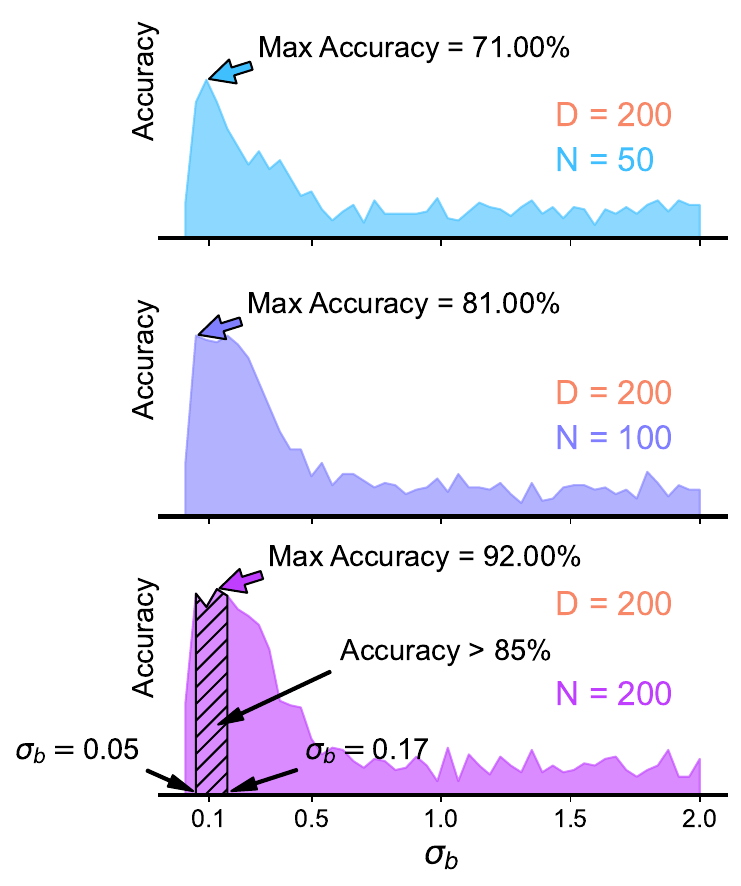}%
\label{fig:acc_starb_lpbf_sub1}} \hfill
\subfloat[]{\includegraphics[width=0.32\textwidth]{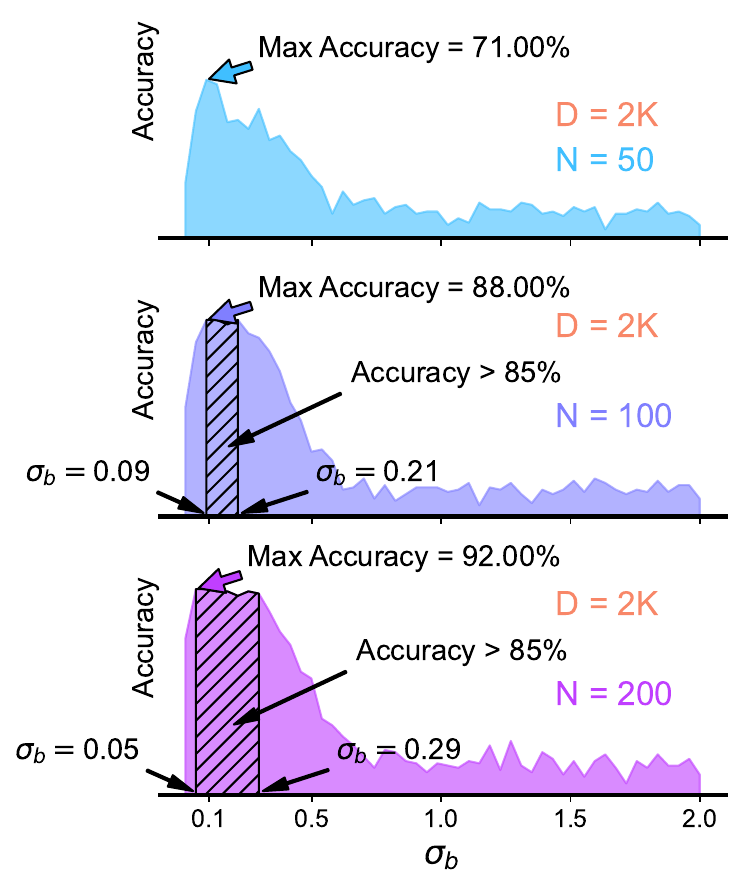}%
\label{fig:acc_starb_lpbf_sub2}} \hfill
\subfloat[]{\includegraphics[width=0.32\textwidth]{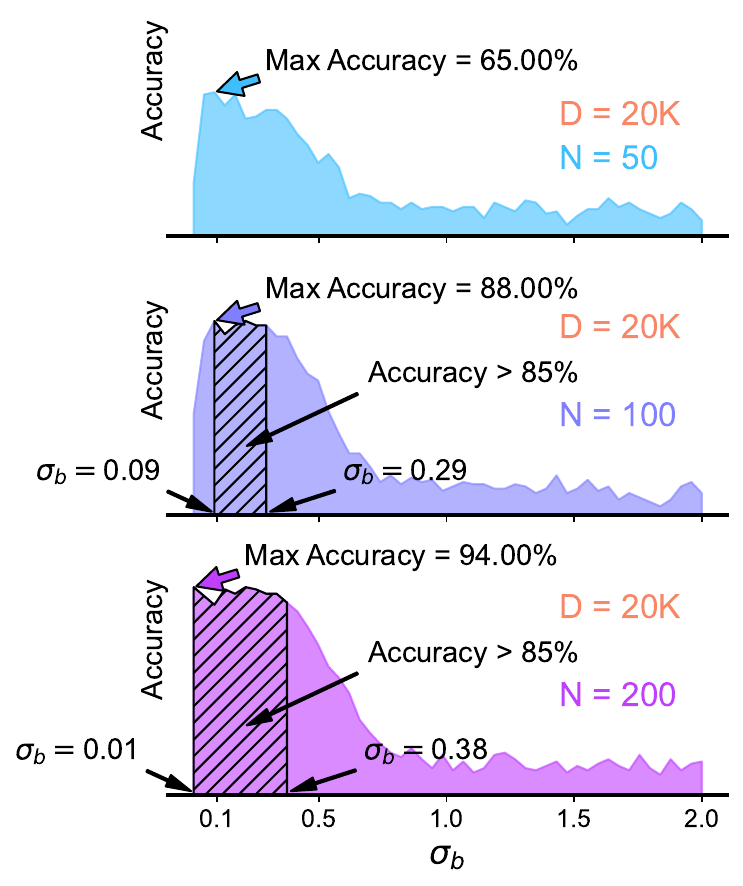}%
\label{fig:acc_starb_lpbf_sub3}}

\caption{Effect of hypervector dimensionality and standard deviation of RFF basis functions on model accuracy for the LPBF task, evaluated with (a) 200, (b) 2K, and (c) 20K training samples.}
\label{fig:acc_N_starb_lpbf}
\end{figure}

Figure~\ref{fig:acc_N_starb_lpbf} illustrates the effect of training set size on the accuracy of the LPBF task. In Figure~\ref{fig:acc_starb_lpbf_sub1}, where the hypervector dimensionality is fixed at \(D = 200\), the number of training samples \(N\) increases from 50 to 200 across different values of \(\sigma_b\). As anticipated, accuracy improves with more training data, rising from \(71\%\) at \(N=50\) to \(92\%\) at \(N=200\). This improvement reflects the benefit of having additional representative patterns for constructing reliable class hypervectors, which enhances the model’s ability to capture visual variations in the LPBF process.  

Unlike the CNC task, where larger hypervector dimensionalities are essential for boosting accuracy, the LPBF task achieves high accuracy even with relatively small hypervectors. This indicates that for LPBF image classification, a dimensionality of \(D = 200\) is already sufficient to capture the discriminative visual information in the data. Increasing \(D\) beyond this level provides limited benefit. For instance, as shown in Figure~\ref{fig:acc_starb_lpbf_sub2}, at \(D = 2{,}000\) the maximum accuracy remains \(92\%\) for \(N=200\), identical to the performance obtained with \(D = 200\). Further expanding the hyperspace to \(D = 20{,}000\), as in Figure~\ref{fig:acc_starb_lpbf_sub3}, yields only a marginal increase to \(94\%\). This 2 percentage point gain comes at the cost of a 100-fold increase in dimensionality and a corresponding rise in inference latency, underscoring the diminishing returns of large hypervector sizes in this task.  

These findings indicate that for the LPBF task, classification accuracy is driven more by the number of training samples than by the dimensionality of the hyperspace. In other words, sample diversity plays a more critical role than representational capacity in capturing the visual patterns necessary for reliable image-based defect detection. This outcome underscores the importance of collecting sufficient representative data in smart manufacturing applications, where expanding hypervector dimensionality yields diminishing returns while increasing computational and energy costs.  

Both the number of training samples \(N\) and the hypervector dimensionality \(D\) shape the sensitivity of performance to the choice of \(\sigma_b\). Selecting \(\sigma_b\) that satisfies a minimum accuracy requirement while remaining within latency and energy budgets is therefore nontrivial. If we set an accuracy floor of \(85\%\), increasing \(D\) broadens the interval of \(\sigma_b\) values that meet this criterion. For example, with \(N = 100\) and \(D = 2{,}000\), only \(\sigma_b \in [0.09,\,0.21]\) attains at least \(85\%\) accuracy, whereas at \(D = 20{,}000\) the valid interval expands to \([0.09,\,0.29]\). This robustness gain must be weighed against the higher inference latency and energy associated with larger \(D\), reinforcing the need for multi-objective, domain-aware tuning.

Increasing the number of training samples similarly reduces the system’s sensitivity to \(\sigma_b\). For a fixed dimensionality of \(D = 20{,}000\), enlarging the training set from \(N = 100\) to \(N = 200\) expands the acceptable range of \(\sigma_b\) from \([0.09,\,0.29]\) to \([0.01,\,0.38]\). This result indicates that greater data availability enhances the robustness of the encoding configuration and reduces uncertainty in parameter selection, thereby simplifying multi-objective tuning under accuracy, latency, and energy constraints.

\begin{figure}
\centering
\includegraphics[width=0.5\textwidth]{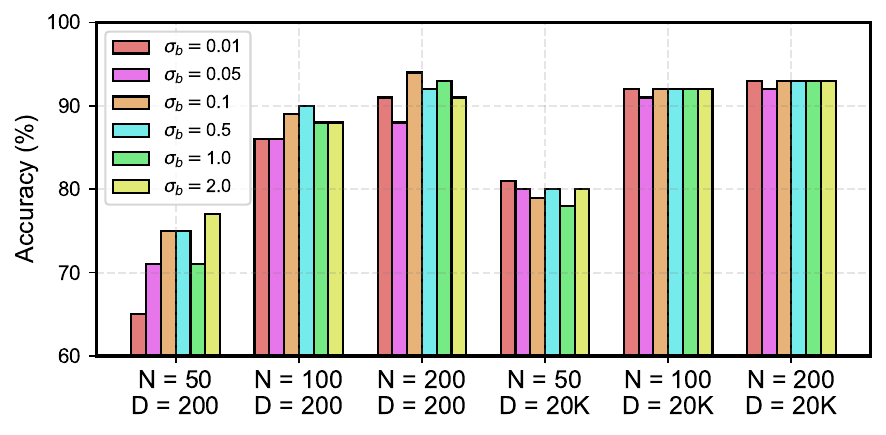}
\caption{Effect of hypervector dimensionality and standard deviation of RP basis functions on classification accuracy for the LPBF task, evaluated across varying training sample sizes.}
\label{fig:acc_rp_barchart}
\end{figure}

Figure~\ref{fig:acc_rp_barchart} reports the accuracy of RP encoding for the LPBF task. For a fixed number of training samples \(N\) and hypervector dimensionality \(D\), variations in \(\sigma_b\) have negligible effect on performance. For example, with \(D = 20{,}000\) and \(N = 100\), accuracy remains at \(92\%\) for all tested values \(\sigma_b \in \{0.01,\,0.1,\,1,\,2\}\). This insensitivity to \(\sigma_b\) indicates that, in the LPBF setting, RP largely decouples accuracy from basis spread, simplifying parameter selection during optimization.

In contrast, varying the number of training samples has a strong effect on accuracy. With \(\sigma_b = 1\) and \(D = 20{,}000\), accuracy rises from \(78\%\) at \(N = 50\) to \(93\%\) at \(N = 200\). The effect is even more pronounced when \(D\) is small. With \(\sigma_b = 0.1\) and \(D = 200\), accuracy increases from \(75\%\) to \(89\%\) to \(94\%\) as \(N\) grows from 50 to 100 to 200, which corresponds to gains of 14 percentage points between \(N = 50\) and \(N = 100\) and a further 5 points between \(N = 100\) and \(N = 200\). These results indicate that, for LPBF with RP encoding, data volume is a primary driver of performance, whereas the basis spread \(\sigma_b\) has minimal impact.

However, when \(D\) is sufficiently large, the HDC model can reach high accuracy with relatively few training samples. For example, at \(D = 20{,}000\), increasing \(N\) from 100 to 200 raises accuracy by only 1\%. This indicates that performance is already saturated near \(N = 100\) and that additional data yields diminishing returns in this regime.

\begin{figure}
\centering
\includegraphics[width=0.5\textwidth]{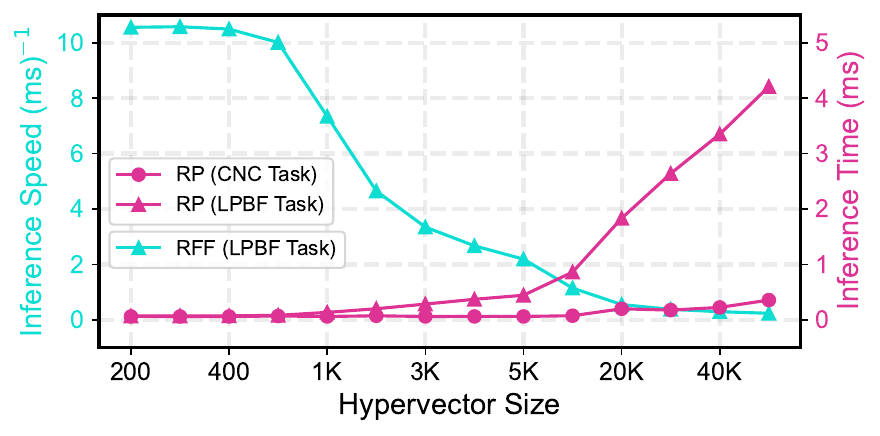}
\caption{Inference time and per-sample inference speed for the CNC and LPBF tasks.}
\label{fig:sensitive_analysis_infer_time}
\end{figure}

Figure~\ref{fig:sensitive_analysis_infer_time} reports the inference time for RP encoding across both tasks. As anticipated, inference latency increases with hypervector dimensionality \(D\). For the CNC task, the measured times are \(0.06\), \(0.07\), \(0.20\), \(0.22\), and \(0.36\)\,\(\mathrm{ms}\) for \(D = 200\), \(2{,}000\), \(20{,}000\), \(40{,}000\), and \(50{,}000\), respectively.

Additionally, the LPBF task consistently exhibits higher inference times than the CNC task, reflecting the larger input feature size of image data relative to multivariate signals. For small \(D\), the gap is modest. At \(D = 200\), LPBF requires \(0.07\,\mathrm{ms}\), only \(0.01\,\mathrm{ms}\) more than CNC. The difference widens markedly as \(D\) grows. At \(D = 20{,}000\), \(40{,}000\), and \(50{,}000\), LPBF incurs \(1.83\,\mathrm{ms}\), \(3.35\,\mathrm{ms}\), and \(4.21\,\mathrm{ms}\), which are approximately \(9.15\times\), \(15.23\times\), and \(11.69\times\) the corresponding CNC times. These results highlight the disproportionate computational burden of high-dimensional encoding for image-based workloads and motivate selecting the smallest \(D\) that satisfies accuracy targets in latency-constrained deployments.

RFF encoding yields inference times that are nearly indistinguishable from RP, indicating that the cosine activation introduces only a marginal latency overhead. For the LPBF task, Figure~\ref{fig:sensitive_analysis_infer_time} reports the inference speed, defined as the inverse of inference time. The observed speeds are \(10.56\), \(4.65\), \(0.55\), \(0.29\), and \(0.24\,\mathrm{ms}^{-1}\) for \(D = 200\), \(2{,}000\), \(20{,}000\), \(40{,}000\), and \(50{,}000\), corresponding to inference times of \(0.09\), \(0.22\), \(1.81\), \(3.39\), and \(4.22\,\mathrm{ms}\), respectively. These results confirm that the choice between RP and RFF has a negligible impact on latency across the tested dimensionalities.

These results confirm that, for the LPBF task, inference with RFF encoding remains highly efficient even at large hypervector dimensionalities. Across all tested values of \(D\), the latency difference between RP and RFF is negligible. For example, at \(D = 50{,}000\) the gap is only \(0.18\%\). This parity reinforces the suitability of RFF for latency-sensitive edge applications, enabling the use of nonlinear features without incurring meaningful runtime cost.

\begin{figure}
\centering

\subfloat[]{%
    \begin{minipage}[b]{0.3\columnwidth}
        \centering
        \includegraphics[width=\linewidth]{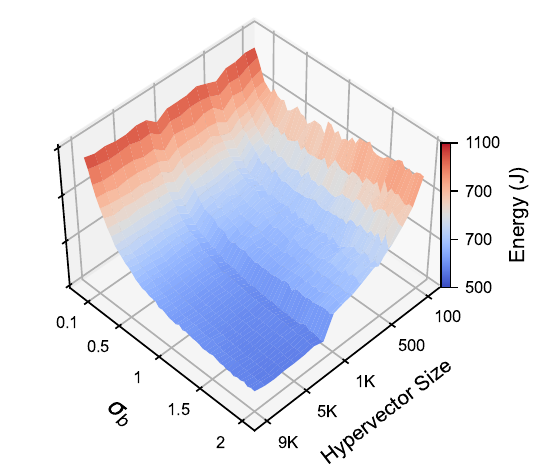}\\[1ex]
        \includegraphics[width=\linewidth]{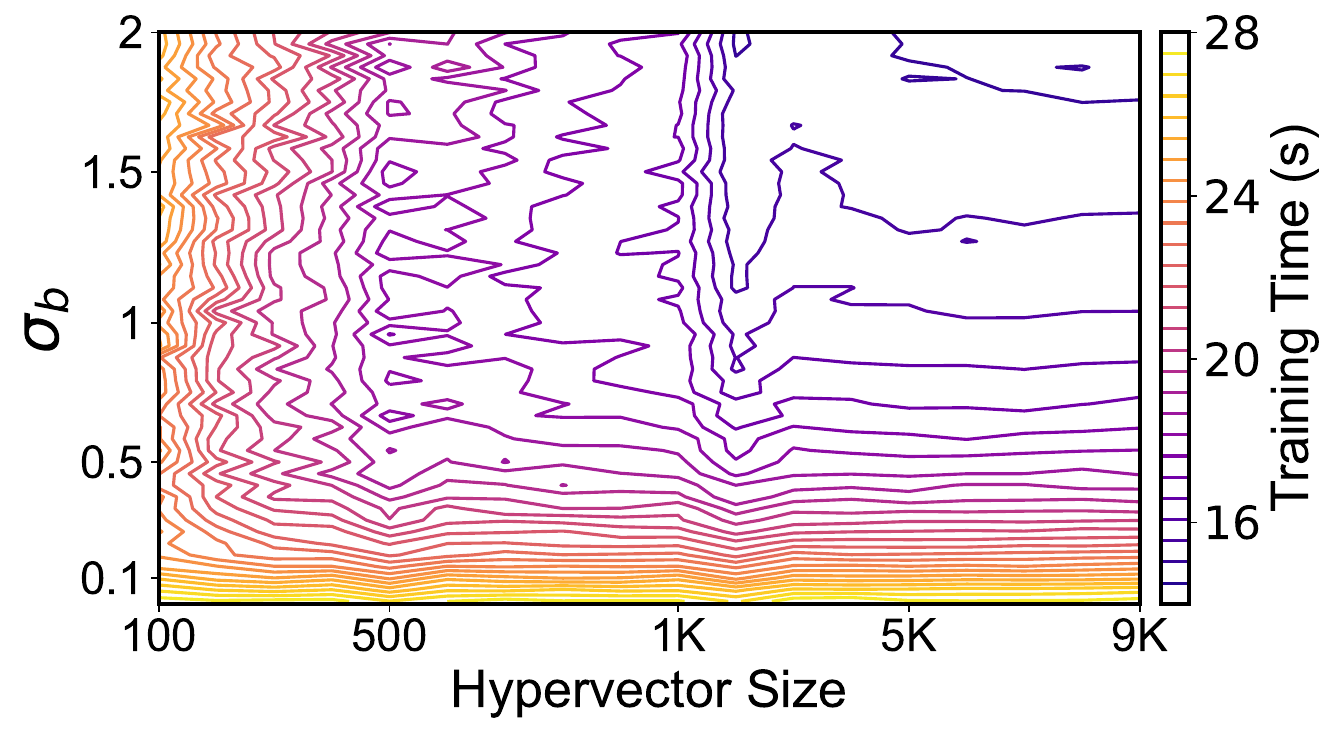}\\[1ex]
        \includegraphics[width=\linewidth]{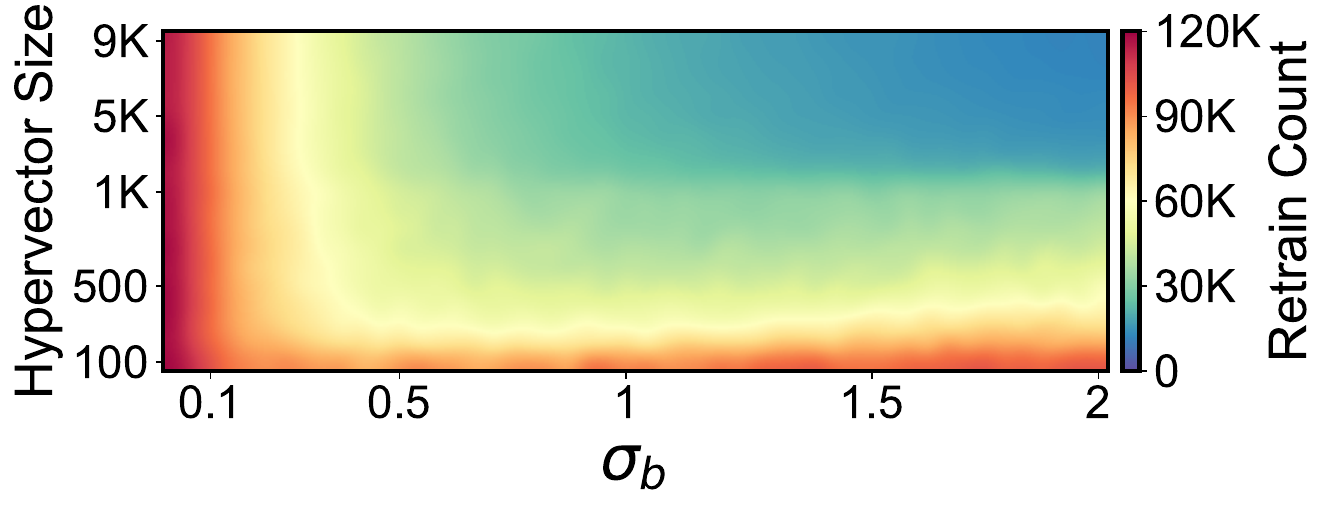}
    \end{minipage}
    \label{fig:time_energy_cnc_task}
}
\subfloat[]{%
    \begin{minipage}[b]{0.3\columnwidth}
        \centering
        \includegraphics[width=\linewidth]{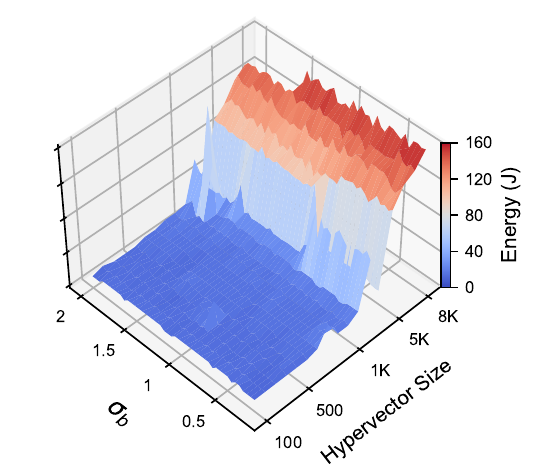}\\[1ex]
        \includegraphics[width=\linewidth]{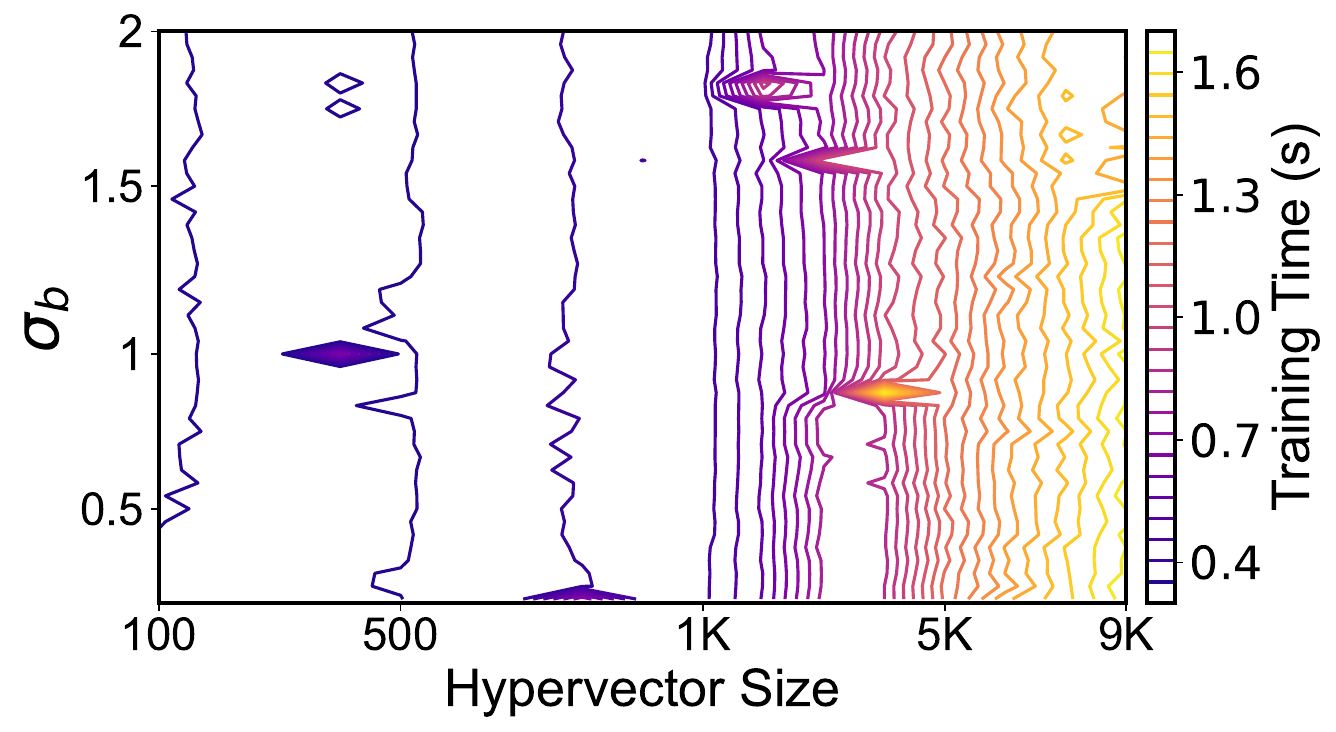}\\[1ex]
        \includegraphics[width=\linewidth]{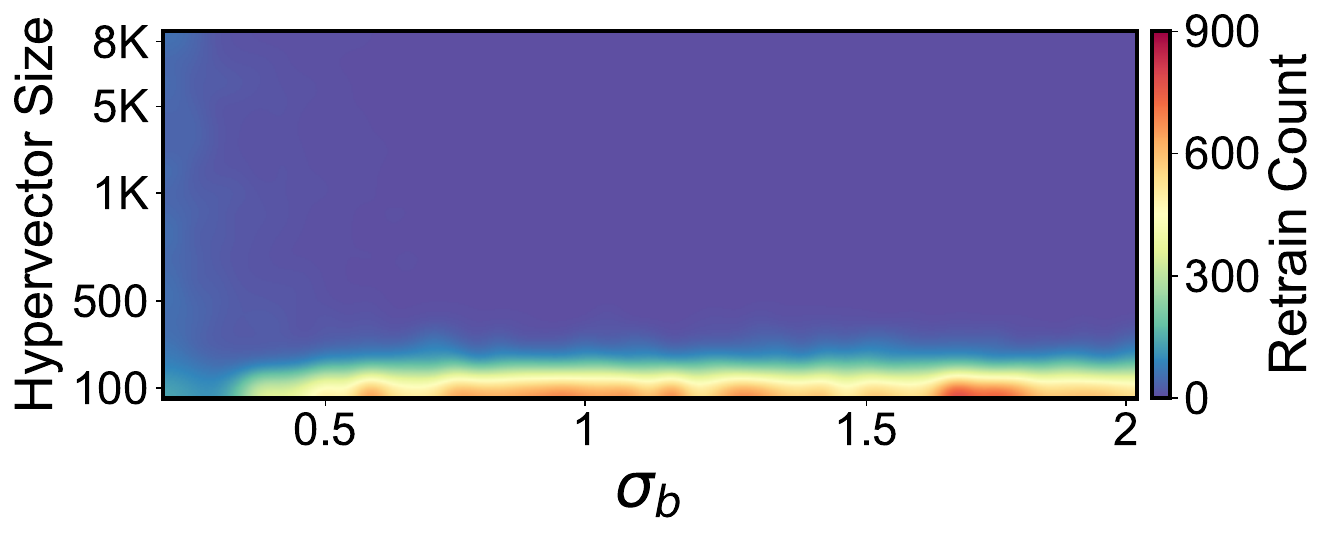}
    \end{minipage}
    \label{fig:time_energy_lpbf_task}
}

\caption{Training time, energy consumption, and number of retrained samples for HDC models using RFF encoding on (a) the CNC task and (b) the LPBF task.}
\label{fig:time_energy_combined_tasks}
\end{figure}

Figure~\ref{fig:time_energy_combined_tasks} summarizes training time, energy consumption, and the number of retraining steps for HDC with RFF encoding. Training time and energy usage are tightly coupled; increases in one are mirrored by the other, indicating that energy is largely driven by computational load. In the CNC task (Figure~\ref{fig:time_energy_cnc_task}), with \(D = 9{,}000\) and \(\sigma_b = 2\), training time is \(14.88\,\mathrm{s}\) and energy consumption is \(581.24\,\mathrm{J}\). In contrast, for \(D = 100\) and \(\sigma_b = 0.01\), both metrics rise to \(27.77\,\mathrm{s}\) and \(1054.04\,\mathrm{J}\) despite the smaller dimensionality. This apparent paradox arises because compact hyperspaces and inclusive encodings reduce separability, increasing misclassifications and the number of updates during retraining. The effect is consistent with the retraining complexity in Eq.~\eqref{eq:hdc_retraining_complexity}, where larger \(P\) directly amplifies training cost.

For the LPBF task (Figure~\ref{fig:time_energy_lpbf_task}), the parameter \(\sigma_b\) exerts minimal influence on training time and energy, whereas both metrics scale primarily with the hypervector dimensionality \(D\). For example, with \((D,\sigma_b)=(100,\,0.21)\) the energy consumption is \(10.96\,\mathrm{J}\) and remains essentially unchanged at \(10.95\,\mathrm{J}\) when \(\sigma_b\) is increased to \(2\). In contrast, increasing \(D\) to \(8{,}000\) raises energy to \(59.30\,\mathrm{J}\). These results indicate that for LPBF, the dominant contributors to training cost are the one pass encoding and prototype construction terms in Eqs.~\eqref{eq:hdc_training_complexity} and \eqref{eq:hdc_retraining_complexity}, which scale with \(D\), while the retraining term is comparatively small. More broadly, the comparison with CNC underscores that sensitivity to \(\sigma_b\) and \(D\) is application dependent, reinforcing the need for domain aware tuning.

The HDC training pipeline comprises three sequential stages: (i) encoding feature vectors into hypervectors, (ii) aggregating the encoded samples to form the initial class hypervectors, and (iii) mistake-driven retraining over multiple epochs. The first two stages are executed once at initialization and their computational cost increases linearly with the hypervector dimensionality \(D\) as given in Eq.~\eqref{eq:hdc_training_complexity}. When the number of misclassified samples is small, these initialization steps dominate wall-clock time and energy. If the initial separation is poor, the retraining term in Eq.~\eqref{eq:hdc_retraining_complexity} grows with the total number of correction steps \(P\) and becomes the primary contributor to training cost.

As shown in Figure~\ref{fig:time_energy_lpbf_task}, training time and energy consumption for the LPBF task are governed primarily by the hypervector dimensionality \(D\), since the number of retraining updates remains below \(1{,}000\) even in the worst case when \(D < 400\). For example, at \(D = 5{,}000\), training time ranges from \(1.12\) to \(1.20\,\mathrm{s}\) and energy ranges from \(92.79\) to \(119.38\,\mathrm{J}\) across all tested values of \(\sigma_b\). In contrast, increasing \(D\) from \(100\) to \(8{,}000\) yields a \(377\%\) rise in training time from \(0.35\) to \(1.67\,\mathrm{s}\) and a \(997\%\) rise in energy from \(14.63\) to \(160.55\,\mathrm{J}\). These results indicate that for LPBF the one pass encoding and prototype initialization terms in Eq.~\eqref{eq:hdc_training_complexity} dominate the overall cost, while the retraining contribution in Eq.~\eqref{eq:hdc_retraining_complexity} is comparatively minor.

Moreover, retraining time and energy depend jointly on the hypervector dimensionality \(D\) and the basis spread \(\sigma_b\). If \(\sigma_b\) fails to project samples into a class-separable region of the hyperspace, or if \(D\) is too small to preserve discriminative information, the initial class prototypes lack separability and misclassifications increase. The resulting rise in correction steps \(P\) drives up both wall-clock time and energy, consistent with the retraining term in Eq.~\eqref{eq:hdc_retraining_complexity}, where cost scales with \(P\) as well as \(D\).

This behavior is evident in the CNC task results in Figure~\ref{fig:time_energy_cnc_task}, where training time, energy usage, and retraining count rise and fall together. When \(D\) is small, costs are high for all values of \(\sigma_b\) because the representational capacity is insufficient. When \(\sigma_b\) is small, the inclusive encoding reduces class separability across all \(D\), which lowers initial accuracy and inflates the number of corrections. These trends match the accuracy degradation observed for inclusive RFF in Figures~\ref{fig:acc_sig_d_3d} and~\ref{fig:acc_N_starb_cnc}, and they are consistent with Eq.~\eqref{eq:hdc_retraining_complexity} where larger \(P\) directly increases training cost. Overall, the CNC signal modality is sensitive to both \(D\) and \(\sigma_b\), reinforcing the need for domain aware selection of encoding parameters.

\subsection{Optimization}
\label{sec:Optimization}

\begin{figure}
\centering

\subfloat[]{\includegraphics[width=0.48\textwidth]{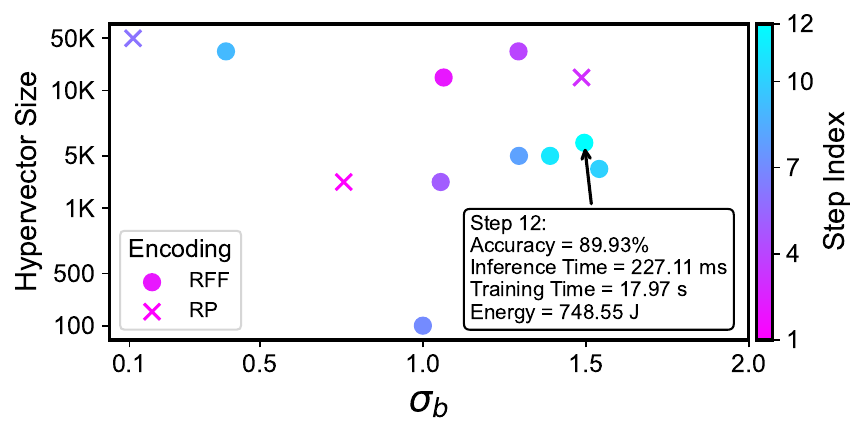}%
\label{fig:optimization_cnc}} 
\subfloat[]{\includegraphics[width=0.48\textwidth]{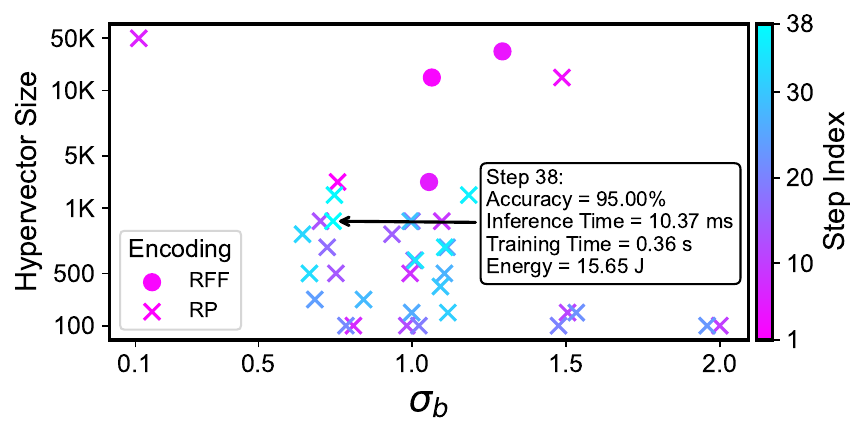}%
\label{fig:optimization_lpbf}}

\caption{Optimization results for the HDC encoding configuration using RFF basis functions for (a) the CNC task and (b) the LPBF task.}
\label{fig:optimization_lpbf_cnc}
\end{figure}

Figure~\ref{fig:optimization_cnc} shows the optimization trajectory for the CNC task. The search began from a randomly selected configuration \(\theta = (t, D, \sigma_b)\). The first trial used \((\mathrm{RP}, 2053, 0.76)\) and produced \(\vec{y} = (\mathrm{Accuracy}=45.83\%,\, \mathrm{Inference\ Time}=199.82~\mathrm{ms},\, \mathrm{Training\ Time}=25.01~\mathrm{s},\, \mathrm{Energy}=1180.78~\mathrm{J})\), which was clearly suboptimal. In response to the low accuracy, subsequent iterations switched to RFF and increased \(D\) beyond \(15{,}000\), consistent with the earlier sensitivity analysis that favored nonlinear encoding and larger hyperspaces for CNC.

To satisfy latency and energy limits, the optimizer briefly reverted to RP while keeping \(D\) large and slightly increasing \(\sigma_b\), which reduced accuracy to \(44.07\%\). It then returned to RFF and selected \((D=31{,}858,\ \sigma_b=1.29)\), achieving \(90.07\%\) accuracy, \(598.32~\mathrm{ms}\) inference time, \(18.15~\mathrm{s}\) training time, and \(934.44~\mathrm{J}\) energy.

In subsequent iterations, the optimizer adjusted \(D\) and \(\sigma_b\) to trade off accuracy and computational cost. At step 7 with \((D=100,\ \sigma_b=1.00)\) the accuracy fell to \(41.57\%\), confirming that a very small hyperspace is inadequate for this task. Increasing \(D\) to \(4{,}116\) while keeping \(\sigma_b=1.29\) restored accuracy to \(90.37\%\). This pattern suggests that \(\sigma_b \approx 1.29\) is a stable encoding choice for CNC, whereas performance is primarily constrained by insufficient dimensionality.

This hypothesis was evaluated by reducing \(\sigma_b\) to \(0.40\) while keeping \(D>30{,}000\), which reduced accuracy to below \(82\%\), confirming strong sensitivity to the encoding spread even at large dimensionality. In the final iterations, the search settled near \(\sigma_b \in [1.39,\,1.54]\) and adjusted \(D\) to balance accuracy and computational cost. The final configuration \((\mathrm{RFF},\, D=5{,}934,\, \sigma_b=1.50)\) achieved \(89.93\%\) accuracy, \(227.11~\mathrm{ms}\) inference time, \(17.97~\mathrm{s}\) training time, and \(748.55~\mathrm{J}\) energy consumption.

Figure~\ref{fig:optimization_lpbf} presents the optimization trajectory for the LPBF task. The search began with a randomly selected configuration \( \theta = (t, D, \sigma_b) = (\mathrm{RP}, 2053, 0.76) \), which already delivered strong performance with \(92.00\%\) accuracy, \(28.32~\mathrm{ms}\) inference time, \(0.56~\mathrm{s}\) training time, and \(23.52~\mathrm{J}\) energy. This behavior is consistent with the sensitivity analysis, where RP achieved high accuracy for LPBF at moderate \(D\) and showed limited dependence on \(\sigma_b\).

In the second step the optimizer switched to RFF with \( (D=15316, \sigma_b=1.06) \) and accuracy collapsed to \(21.00\%\). The search immediately reverted to RP, which restored accuracy to \(93.00\%\). Additional trials with RFF repeatedly produced accuracies below \(20\%\), so the remaining 33 evaluations stayed with RP. Throughout the trajectory, RP consistently delivered accuracy above \(90\%\) with only mild variation across \(D\) and \(\sigma_b\), confirming the LPBF preference for linear projection and its low sensitivity to basis variance.

Several configurations were evaluated to probe the trade-off between performance and efficiency. At step 7 the setting \(D=100\) with \(\sigma_b=0.81\) achieved \(91.00\%\) accuracy with inference time \(5.80\,\mathrm{ms}\), training time \(0.34\,\mathrm{s}\), and energy \(12.82\,\mathrm{J}\). This shows that a compact hyperspace can sustain high accuracy while operating in a very low-latency regime. Keeping \(D=100\) and increasing \(\sigma_b\) to \(2.00\) reduced accuracy to \(87.00\%\), highlighting the importance of selecting an appropriate encoding regime, where inclusive settings are preferable to exclusive ones for this task at small \(D\).

A second configuration with \(D=717\) and \(\sigma_b=0.64\) reached \(92.00\%\) accuracy with inference time \(9.82\,\mathrm{ms}\) and energy \(11.79\,\mathrm{J}\). Compared with the \(D=100,\ \sigma_b=0.81\) setting, this yields a \(+1.00\) percentage point gain in accuracy and a \(1.03\,\mathrm{J}\) reduction in energy, at the cost of higher latency \((5.80\,\mathrm{ms}\rightarrow 9.82\,\mathrm{ms})\). This illustrates a Pareto trade-off in which a moderate increase in \(D\) combined with a tuned \(\sigma_b\) improves accuracy and lowers energy while keeping inference within a single-digit millisecond regime.

The final configuration \(D=837,\ \sigma_b=0.74\) achieved \(95.00\%\) accuracy with inference time \(10.37\,\mathrm{ms}\), training time \(0.36\,\mathrm{s}\), and energy \(15.65\,\mathrm{J}\). This point lies near the Pareto front, offering a balanced trade-off among accuracy, latency, and energy, and is therefore a strong candidate for deployment in smart manufacturing settings.

\subsection{Benchmark}
\label{sec:Benchmark}

\begin{figure}
\centering

\subfloat[]{\includegraphics[width=0.25\columnwidth]{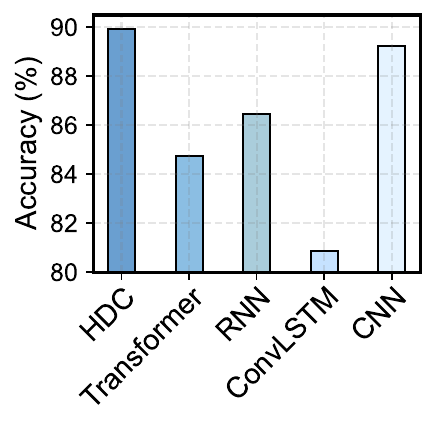}%
\label{fig:bench_acc_cnc}} 
\subfloat[]{\includegraphics[width=0.25\columnwidth]{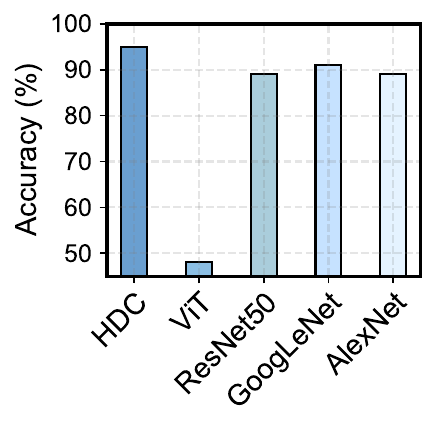}%
\label{fig:bench_acc_lpbf}}

\caption{Accuracy comparison between HDC and deep learning models. (a) CNC task: Transformer, RNN, ConvLSTM, and CNN. (b) LPBF task: Vision Transformer (ViT), ResNet-50, GoogLeNet, and AlexNet.}
\label{fig:bench_acc}
\end{figure}

Figure~\ref{fig:bench_acc} presents the accuracy comparison between the optimized HDC models and various benchmark architectures. For the CNC task, shown in Figure~\ref{fig:bench_acc_cnc}, HDC achieves an accuracy of 89.93\%, outperforming Transformer (84.73\%), RNN (86.46\%), and ConvLSTM (80.86\%), and slightly exceeding CNN (89.23\%). These results demonstrate that HDC offers competitive or slightly better performance than both sequence-based and convolutional models in signal-based classification tasks.

For the LPBF task, Figure~\ref{fig:bench_acc_lpbf} compares HDC to several standard vision models. The optimized HDC model achieves 95\% accuracy, surpassing ViT (48\%), ResNet50 (89\%), GoogLeNet (91\%), and AlexNet (89\%) by 47\%, 6\%, 4\%, and 6\%, respectively. These findings highlight HDC's superlative effectiveness in handling image-based classification tasks typical in additive manufacturing and smart manufacturing environments. Notably, HDC achieves this performance without relying on complex attention mechanisms or deep feature hierarchies, indicating its ability to efficiently capture the correlation between input patterns and class labels across diverse sensing modalities.

\begin{figure}
\centering

\subfloat[]{\includegraphics[width=0.48\textwidth]{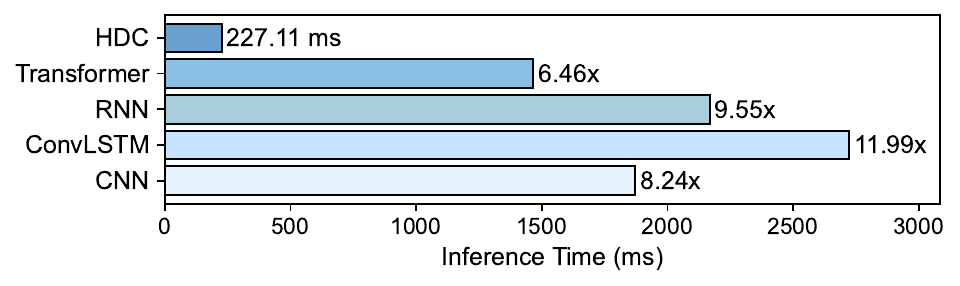}%
\label{fig:bench_infertime_cnc}} 
\subfloat[]{\includegraphics[width=0.48\textwidth]{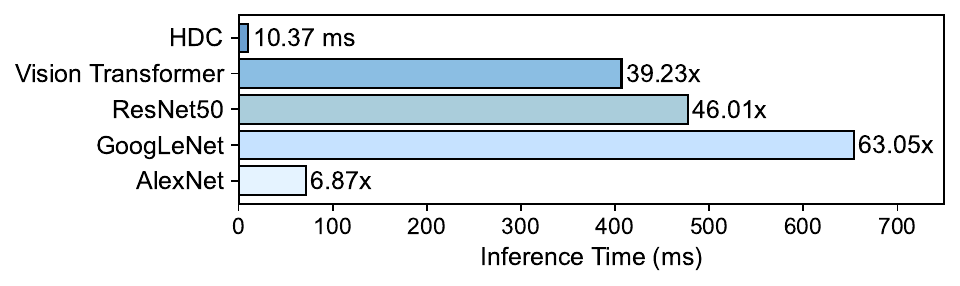}%
\label{fig:bench_infertime_lpbf}}

\subfloat[]{\includegraphics[width=0.48\textwidth]{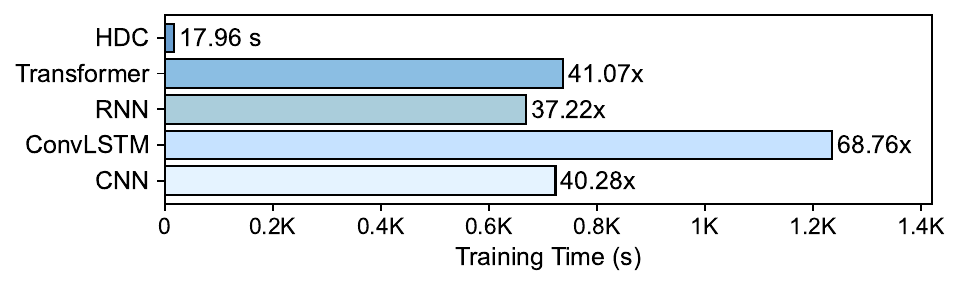}%
\label{fig:bench_traintime_cnc}} 
\subfloat[]{\includegraphics[width=0.48\textwidth]{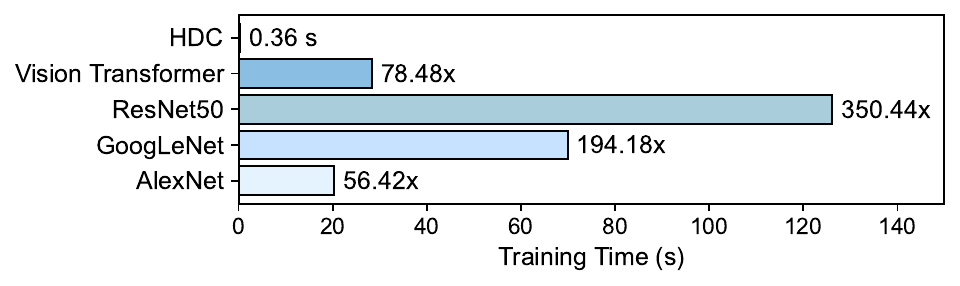}%
\label{fig:bench_traintime_lpbf}}

\caption{Comparison of HDC with deep learning models in terms of inference and training time. (a) Inference time on the CNC task against Transformer, RNN, ConvLSTM, and CNN. (b) Inference time on the LPBF task against Vision Transformer (ViT), ResNet-50, GoogLeNet, and AlexNet. (c) Training time on the CNC task. (d) Training time on the LPBF task.}
\label{fig:bench_train_infertime}
\end{figure}

As the primary objective in smart manufacturing quality control applications is to enable real-time decision-making, inference latency plays a critical role. Figure~\ref{fig:bench_infertime_cnc} illustrates that the optimized HDC model achieves an inference time of 227.11~ms for the CNC task. In comparison, the inference times for the Transformer, RNN, ConvLSTM, and CNN models are 6.46$\times$, 9.55$\times$, 11.99$\times$, and 8.24$\times$ higher, respectively. These results highlight HDC’s efficiency in signal-based tasks where rapid predictions are essential.

Similarly, Figure~\ref{fig:bench_infertime_lpbf} shows the inference time results for the LPBF task. The HDC model achieves an inference time of 10.37~ms, significantly outperforming its vision-based counterparts. The inference times for ResNet50, GoogLeNet, and AlexNet are all at least 71.26~ms, over 6.87$\times$ slower than HDC. ViT is particularly computationally demanding, with an inference time 39.23$\times$ greater than that of HDC. This substantial latency is primarily attributed to ViT’s reliance on self-attention mechanisms and large fully connected layers.

Overall, these findings confirm that HDC offers a substantial inference speed advantage over traditional deep learning models, making it highly suitable for low-latency, resource-constrained environments.

Training latency and computational efficiency are critical considerations for edge deployment in smart manufacturing. Figure~\ref{fig:bench_traintime_cnc} shows that the HDC model trains in 17.96\,s on the CNC task. By comparison, the Transformer, RNN, ConvLSTM, and CNN require 41.07$\times$, 37.22$\times$, 68.76$\times$, and 40.28$\times$ longer, respectively. This demonstrates the training-time advantage of HDC for signal-based quality monitoring.

Similarly, Figure~\ref{fig:bench_traintime_lpbf} reports the training times for the LPBF task. HDC completes training in just 0.36~seconds, whereas ResNet50, GoogLeNet, and AlexNet require at least 56.42$\times$ more time. ViT, due to the complexity of training attention coefficients and large feedforward layers, takes 78.48$\times$ longer than HDC.

\begin{figure}
\centering

\subfloat[]{\includegraphics[width=0.35\columnwidth]{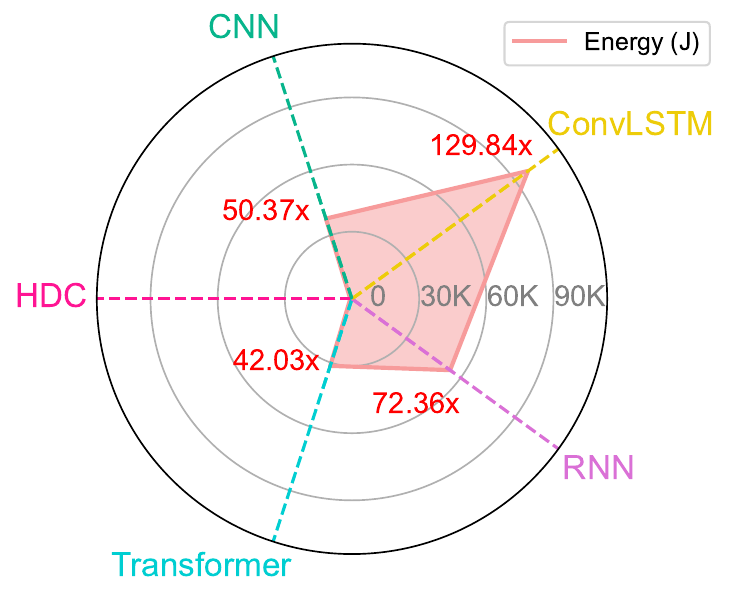}%
\label{fig:bench_energy_cnc}} 
\subfloat[]{\includegraphics[width=0.35\columnwidth]{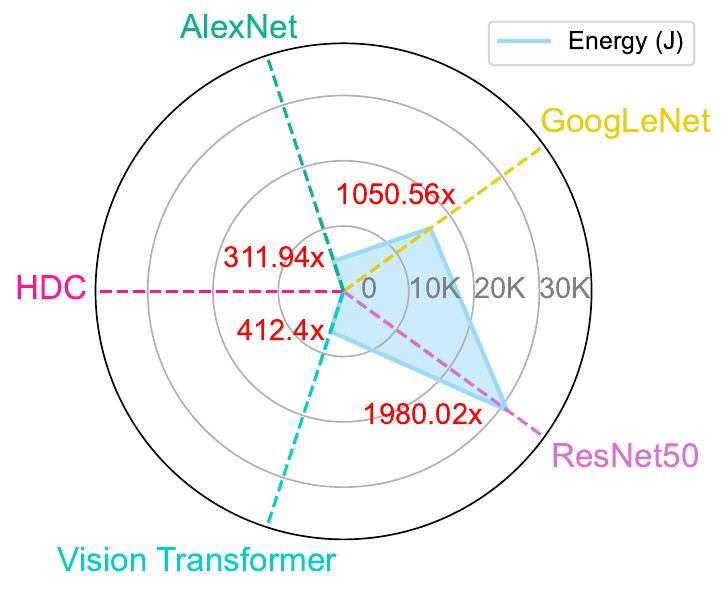}%
\label{fig:bench_energy_lpbf}}

\caption{Training energy consumption comparison between HDC and deep learning models. (a) CNC task: Transformer, RNN, ConvLSTM, and CNN. (b) LPBF task: Vision Transformer (ViT), ResNet-50, GoogLeNet, and AlexNet.}
\label{fig:bench_energy}
\end{figure}

Figure~\ref{fig:bench_energy} illustrates the energy consumption during model training. For the CNC task (Figure~\ref{fig:bench_energy_cnc}), HDC consumes only 748.54~Joules, while Transformer, RNN, ConvLSTM, and CNN require 31{,}461.99, 54{,}167.39, 97{,}190.62, and 37{,}704.62~Joules, respectively, representing energy usage at least 42.03$\times$ higher than HDC. 

Similarly, for the LPBF task (Figure~\ref{fig:bench_energy_lpbf}), HDC consumes just 15.65~Joules. In contrast, ViT, ResNet50, GoogLeNet, and AlexNet consume 6{,}454.09, 30{,}987.32, 16{,}441.31, and 4{,}881.87~Joules, respectively, amounting to 412.4$\times$, 1980.02$\times$, 1050.56$\times$, and 311.94$\times$ greater energy consumption compared to HDC.

This efficiency stems from HDC's training process, which leverages fixed random projections and additive updates without iterative weight optimization or attention learning. In contrast, Transformer-based and neural network-based models require numerous epochs of gradient-based updates to learn internal representations, leading to substantially higher energy costs.

\section{Discussion}
\label{sec:Discussion and Limitations}

This study set out to test a common assumption in the HDC literature that the qualitative structure of trade-offs between parameters and performance is universal across applications. Our results show that the assumption does not always hold. The CNC task favors RFF with an exclusive spread, while the LPBF task favors RP with an inclusive spread and modest dimensionality. The same hyperparameters can therefore induce different profiles for accuracy, latency, and energy when the sensing modality and task demands change. This observation reframes HDC design in smart manufacturing as a domain-aware problem rather than a single recipe that fits all cases.

The analytical results provide a mechanism for the empirical findings. Inference scales with \(\mathcal{O}(JD + LD)\) as shown in Eq.~\eqref{eq:hdc_inference_complexity}, and training scales with \(\mathcal{O}(ND(J+1))\). Larger \(D\) increases representational capacity but also increases the cost per operation. Retraining adds a workload proportional to \(P\), where \(P\) is itself a coupled function of the encoding type \(t\), the projection variance \(\sigma_b\), the feature dimension \(J\), the number of classes \(L\), and the structure of the data. These interactions create non-monotonic trade-offs. Some effects, such as the growth of inference time with \(D\), are predictable from the complexity terms, while others, such as the number of corrections \(P\), depend on how a particular encoding separates classes in a given domain and must be measured empirically.

Together, the theory and experiments motivate a practical design stance for smart manufacturing. First, treat HDC configuration as a multi-input and multi-output problem with accuracy, inference time, training time, and energy as concurrent objectives. Second, use domain signals to choose between linear and nonlinear encoders. Temporal signals with subtle phase interactions benefit from RFF and larger \(\sigma_b\), whereas spatially structured images can succeed with RP and small to moderate \(D\). Third, match \(D\) to the minimum that achieves the target accuracy under the application latency budget, since Eq.~\eqref{eq:hdc_inference_complexity} implies that excess dimensionality yields higher cost without guaranteed gains. Fourth, view data volume as a control knob that can widen feasible regions in \((D,\sigma_b)\) space, which improves robustness during deployment.

These findings also clarify the role of optimization. The goal is not to introduce a new optimizer, but to recognize that no simple closed-form mapping exists from \((t,D,\sigma_b,J,L,N)\) to the joint objectives. A sample-efficient search, such as Bayesian optimization, is therefore a reasonable instrument for discovering feasible operating points under domain constraints. The novelty lies in establishing that the feasible set shifts qualitatively with the application, and in showing that a tuned HDC can outperform deep baselines on latency and energy while maintaining competitive accuracy in real manufacturing tasks.

There are limitations. The study considers two representative tasks, one signal based and one image based, which do not span the full diversity of manufacturing processes. The search space focused on \(t\), \(D\), and \(\sigma_b\). Other HDC choices, such as permutation stride, sparsity, channel inclusion rules, and alternative similarity metrics, were not explored. Energy and timing were measured on a general-purpose platform that provides precise instrumentation, whereas ultimate deployment targets will include embedded and accelerator platforms, and porting the tuned configurations to such hardware remains future work. Extreme class imbalance and severe data scarcity were not the primary focus and deserve a dedicated study. Finally, we did not analyze robustness under distribution shift, which is common on factory floors due to tool wear and sensor drift.

Future work should extend the analysis in three directions. First, enlarge the parameter set and derive tighter bounds that connect separability in the encoded space to expected corrections \(P\). Second, develop online budget-aware tuning that adapts \((t,D,\sigma_b)\) in response to drift while keeping latency within fixed limits. Third, build cross-domain priors that warm start the search yet remain conservative about transfer, reflecting the domain dependence revealed here.

\section{Conclusion and Future Works}
\label{sec:Conclusion and Future Works}

This paper examined a widely held assumption in the HDC literature that the qualitative relation between parameters and performance is universal. Evidence from two smart manufacturing tasks shows that this assumption does not always hold. Signal based CNC classification favored RFF with an exclusive spread, while image based LPBF classification favored RP with an inclusive spread and modest dimensionality. The same hyperparameters produced different accuracy, latency, and energy profiles once the sensing modality and task demands changed. The central implication is that HDC must be configured with domain aware encoders rather than relying on a single recipe.

The analytical results clarified why such divergence appears. Inference and training scale linearly with feature size and dimensionality, yet the number of correction steps during retraining depends on how the chosen encoder separates classes in a given domain. This coupling creates non-monotonic trade-offs that are not captured by a simple closed form. Treating HDC design as a multi-input and multi-output optimization is therefore necessary. A simple sample-efficient search was sufficient to identify feasible operating points under domain constraints, and tuned HDC models outperformed deep baselines on latency and energy while maintaining competitive accuracy.

The experiments provide practical guidance for the engineering deployment in smart manufacturing. Choose between linear and non-linear encoders using domain cues. Match the hypervector dimensionality to the smallest value that meets the target accuracy under the latency budget. Increase training data when possible to widen the feasible region for projection variance, which improves robustness to parameter choices. Avoid transferring heuristics across domains without validation.

Several limitations suggest concrete next steps. Future work will broaden the study to additional manufacturing processes and sensing modalities, incorporate a larger set of HDC controls such as permutation stride, sparsity, and similarity metrics, and develop online budget aware tuning that adapts encoder type, dimensionality, and projection variance under fixed latency and energy limits. Porting the tuned configurations to embedded and accelerator platforms and measuring on device energy will close the loop to deployment. Finally, tighter bounds that link separability in the encoded space to the expected number of corrections, together with evaluations under class imbalance, distribution shift, and continual learning, will further strengthen the design rules revealed here.

\section*{Acknowledgments}
This work was supported by the National Science Foundation, United States [grant numbers 2434519, 2434385, 2146062, and 2001081]; the Department of Energy [grant number DE-EE0011029]; and UConn Startup Funding. The authors gratefully acknowledge the contributions of the National Institute of Standards and Technology, particularly Dr. Brandon Lane, for providing the Laser Powder Bed Fusion dataset used in this study. They also acknowledge the support of the Connecticut Center for Advanced Technology, especially Nasir Mannan, for providing the Computer Numerical Control dataset.

\bibliographystyle{unsrt}  
\bibliography{references}  
\end{document}